\documentclass[journal]{IEEEtai}
\usepackage[colorlinks,urlcolor=blue,linkcolor=blue,citecolor=blue]{hyperref}
\usepackage{xcolor,array}
\usepackage{graphicx}
\let\labelindent\relax
\usepackage{enumitem}
\usepackage{tabularx}
\usepackage{booktabs}
\usepackage{multirow}
\usepackage{amsmath,amssymb,amsfonts}
\usepackage{listings}
\usepackage{pgfplots}
\usetikzlibrary{arrows.meta,shapes.geometric}
\usepackage{cite}
\usepackage{subcaption}
\pgfplotsset{compat=1.18}
\usepgfplotslibrary{colormaps,statistics,groupplots}
\pgfplotsset{
    colormap={greengradient}{
        rgb255=(238,247,242)
        rgb255=(10,67,36)
    },
}

\definecolor{successGreen}{RGB}{0,128,0}

\newcommand{\omini}{\texttt{$\mu$-O4}}
\newcommand{\gem}{\texttt{$\varphi$-G3}}

\newif\ifshowrevisions
\showrevisionsfalse
\definecolor{revisionGreen}{RGB}{0,100,60}
\newcommand{\revision}[1]{\ifshowrevisions{\color{revisionGreen}#1}\else#1\fi}
\newcommand{\reviewchange}[1]{#1}
\newcommand{\promptlocation}{Appendix~\ref{app:waterfilling_initial_prompt}}
\newcommand{\comparisonlocation}{Appendix~\ref{app:sum_rate_solution_comparison}}
\hypersetup{pdfauthor={Lingyun Guo and Ali Bereyhi}}
\begin{document}
\bstctlcite{IEEEexample:BSTcontrol}
\title{Understanding Large Language Models as Numerical Optimizers}
\author{Lingyun Guo and Ali Bereyhi, \textit{Senior Member, IEEE}%
\thanks{Lingyun Guo and Ali Bereyhi are with the Edward S. Rogers Sr. Department of Electrical and Computer Engineering (ECE), University of Toronto, Toronto, ON, Canada; emails: \texttt{lingyun.guo@mail.utoronto.ca} and \texttt{ali.bereyhi@utoronto.ca}. }%
% \thanks{Ali Bereyhi is with the Department of Electrical and Computer Engineering, University of Toronto, Toronto, ON, Canada.}%
\thanks{Ali Bereyhi is Faculty Affiliate at the Schwartz Reisman Institute (SRI) for Technology and Society.}}
\markboth{}%IEEE Transactions on Artificial Intelligence}
{Guo and Bereyhi: Understanding Large Language Models as Numerical Optimizers}
\maketitle
\begin{abstract}
Large language models (LLMs) have emerged as potential tools for numerical optimization. Despite promising results, their intrinsic optimization capabilities and limitations remain poorly understood, particularly in high dimensions. This paper investigates the ability of LLMs to exploit mathematical problem structures and generate feasible, near-optimal solutions. We develop an evaluation framework that distinguishes solution quality from execution failures, including numerical inaccuracies, constraint violations, and output-format errors. Using representative problems from convex optimization, fractional programming, and constrained projection, we assess LLM performance against deterministic reference solvers across different dimensions. Our results reveal that LLMs can capture useful optimization structures and produce near-optimal solutions, even when their outputs fail to satisfy numerical or formatting requirements. However, reliability deteriorates with increasing problem dimension and depends strongly on the underlying optimization structure. To further investigate the limitations, we deploy an iterative refinement mechanism that provides targeted feedback on feasibility and solution quality. The results demonstrate that iterative refinement substantially improves reliability and extends the range of problem dimensions over which LLMs produce valid solutions. Nevertheless, persistent numerical and structural errors limit their applicability as standalone optimizers. These findings provide insights into the capacity and limits of LLM-based optimization and motivate hybrid approaches that combine LLM reasoning with deterministic numerical verification.
\end{abstract}

\begin{IEEEImpStatement}
Large language models (LLMs) are increasingly being considered for optimization and automated decision-making in engineering and other computational applications. Deploying these models without understanding their numerical limitations may lead to infeasible or unreliable decisions, particularly in systems governed by strict operational constraints. This work provides empirical insights into the capabilities and limitations of LLMs as numerical optimizers. It depicts that the ability of LLMs to identify useful structures does not necessarily translate into accurate numerical execution. By characterizing failure modes and investigating the impact of iterative feedback, this study highlights the importance of independent feasibility verification and numerical validation. The findings can inform the development of more reliable LLM-aided optimization systems, in which LLMs support problem interpretation and solution generation while deterministic algorithms enforce mathematical rigour. Such hybrid approaches may facilitate the responsible integration of LLMs into engineering optimization and automated decision-making workflows.
\end{IEEEImpStatement}


\section{Introduction}

Recent advances in large language models (LLMs) have demonstrated their ability to interpret natural language context, take feedback, and generate solutions through reasoning \cite{LLM_FewShot,LLMs_RLHF,CoTh,sun2025survey}. These capabilities motivate the exploration of LLMs as numerical optimizers for different tasks, especially in applications such as resource allocation and scheduling \cite{llm_case_study,scheduling_llm,Survey}.

Existing work has studied LLM-based optimization methods in various contexts. Yang et al. \cite{llm_as_optimizers} consider optimization tasks described in natural language, including linear regression, the traveling salesman problem, and prompt optimization for task accuracy, and propose the \textit{optimization by prompting (OPRO)} scheme, which iteratively generates candidate solutions using the guidance of an external evaluator. The \textit{LLM-driven evolutionary algorithm (LMEA)} approach applies mutation and crossover operations on high-quality candidates to search for better solutions \cite{llm_as_evolutionary_optimizers}. Closely related lines of work have further explored learning-based approaches instead of direct LLM prompting \cite{li2016learning,OptSun,JMLR_LearnOpt,llmopt}. A recent example is the \textit{unified learning-based optimization (LLMOPT)} framework in \cite{llmopt} that  proposes fine-tuning LLMs to represent and solve optimization problems more reliably. LLMOPT also introduces mechanisms for \textit{self-correction} and improved output validation.

\subsection{Motivation: Reliability of LLM-based Optimizers}
The mentioned studies suggest that LLMs can serve as a useful component in optimization pipelines. Nevertheless, the available lines of work focus mostly on developing new designs that improve the solution inferred from a given context and pay less attention to understanding the intrinsic optimization capabilities of LLMs. In numerical optimization, however, baseline algorithms rely on the \textit{problem structure} deterministically and exploit it to determine the solution, e.g., water-filling for parallel-channel power allocation, Dinkelbach iterations for fractional programming, and the alternating direction method of multipliers (ADMM)  for constrained optimization \cite{boyd2004convex,fp_part1,fp_part2,dinkelbach1967,admm}. This enables the algorithms to guarantee satisfaction of feasibility constraints and numerical precision requirements. Although prior work demonstrates the capability of LLM-based pipelines for inferring the solution, it fails to understand the underlying intrinsic process. The lack of this knowledge, restricts the reliability and efficiency of these optimizers, especially in high-dimensional and high-precision settings. In particular, there is limited empirical evidence on how LLM performance scales with problem dimension or whether LLMs effectively exploit the underlying problem structure throughout the reasoning tree. Existing studies do not clearly distinguish poor decisions from execution failures. This lack of understanding makes the guaranties for the feasibility, precision, and optimality gap of the output unclear. This work develops a framework that addresses this research gap.

\subsection{Contributions and Organization}
We study the intrinsic behaviour of LLMs, when deployed as numerical optimizers. Our contributions are as follows.
\begin{enumerate}
    \item We propose a framework that evaluates how \reviewchange{an} LLM exploits the problem structure throughout its optimization pipeline. The framework uses a loop to assess the main feature extracted by the LLM and distinguish poor in-context learning from execution errors.
    \item Motivated by recent findings on reasoning procedure in LLMs \cite{lin2025zebralogic,GSM}, we propose an iterative refinement protocol for numerical optimization via LLMs. In this protocol, the LLMs receive an assessment about the structure of the solutions they previously produced and are prompted to propose an improvement. This protocol enables LLMs to correct \textit{execution} errors and improve their solutions over multiple attempts, which as a result provides means to evaluate the understanding of problem structure by the LLM. 
    \item We study three different optimization families, namely convex optimization, fractional programming, and optimization problems with mixed \reviewchange{equality} and inequality constraints. For each family, we evaluate whether LLMs exploit the problem structure in its reasoning chain and investigate the feasibility and optimality of inferred solutions by comparing the results with that of deterministic optimal solvers. Our findings suggest that LLMs use the problem context to estimate the structure of the underlying problem and utilize it to infer the solution.
    \item We study the scalability of LLM-based optimization for the considered three problem families. For each family, we characterize how the reliability and accuracy of the inferred solution scales with dimension and identify practical failure modes. The results are compared to deterministic reference solvers in terms of feasibility, optimality, and execution time. The results are consistent with our earlier findings: in problems with known structure, LLM-based optimization can better scale, while in problems with unknown algorithmic solution, the evaluation loop starts to report failure at lower dimensions. This finding suggests that the underlying structure actively contributes to the inferred solution by LLM. 
\end{enumerate}

The remainder of this manuscript is organized as follows: Section~\ref{sec:related} reviews the background and related work on LLM-based optimization. We present our proposed framework in Section~\ref{sec:framework} and elaborate the evaluation metrics. Experimental settings are presented in Section~\ref{sec:experiments}, where we report our findings. The results and their implications for LLM-based optimization are discussed in Section~\ref{sec:discussions}. Finally, the paper is concluded in Section~\ref{sec:conclusions}.

\section{Related Work}
\label{sec:related}
The work on LLM-based optimization can be categorized into two main directions: ($i$) design of LLM-assisted optimizers, and ($ii$) evaluation of LLM-based optimization pipelines. In the sequel, we go over these two directions. 
\subsection{LLMs as Optimizers}
Classical LLM-based optimizers use prompting methods to infer a solution for the underlying optimization. A representative approach that deploys LLMs as general-purpose optimizers is OPRO \cite{llm_as_optimizers}. The approach solves problems in an iterative manner. An LLM first proposes candidate solutions, and an external scorer evaluates them. The best candidates are then summarized and returned to the LLM to guide the next round of proposals. This black-box workflow is attractive for problems in which ($i$) gradients are unavailable, ($ii$) the objective is only accessible through simulation, and ($iii$) the constraints are difficult to encode in standard solvers. An alternative, called LMEA, approach is proposed in \cite{llm_as_evolutionary_optimizers}. LMEA applies mutation and crossover operations to diversify high-quality candidate solutions and combine them. The experiments in \cite{llm_as_evolutionary_optimizers} depict that LMEA can achieve competitive performance on combinatorial benchmarks. 

A prior line of work on the use of learning models for optimization considers learning-based approaches to train models as optimizers \cite{li2016learning,OptSun,JMLR_LearnOpt,llmopt}.  Early studies in this direction focus on explicit training of learning models as solvers; see for example \cite{li2016learning,OptSun}. With advances in LLMs, this approach has been extended to LLM-based optimization. The study in \cite{llmopt} proposes the LLMOPT framework that uses textual problem information to fine-tune LLMs as optimization solvers. The post-training in LLMOPT yields  more reliability, as the LLM learns to attend to the properties of the underlying problem. LLMOPT further includes mechanisms for self-correction and improved output validity. The study in \cite{llm_case_study} \reviewchange{develops} a retrieval-based approach for the specific problem of power allocation in wireless networks. In the proposed approach, an LLM is fed with numerical representations of the network state and selects the best allocations from a quantized action set. Selections are evaluated using a task-specific reward. The solution to each sample problem is then stored in an experience pool. For subsequent decisions, examples are retrieved from this pool and fed back to LLM as prompts to adapt its behaviour to the underlying optimization problem.

Although findings in the former studies suggest that LLMs can serve as proposal generators in optimization pipelines, they highlight various practical challenges when dealing with strict feasibility requirements, e.g., non-negativity of solutions, \reviewchange{equality} budget constraints, and fixed dimensionality \cite{llmopt,llm_case_study}. The limits of LLM-based optimizers in these challenging settings come from two different sources: ($i$) failure in grasping the problem structure, and ($ii$) errors in execution. While the latter can be refined through feedback loops in the pipeline, the former seems to be more restrictive. This work develops a framework to distinguish between these sources and characterize the limits of LLM-based optimizers.

\subsection{LLM Capability Evaluation}

Recent work has developed basic schemes to evaluate the performance of LLMs as optimizers in applications such as code generation and mathematical problem solving \cite{optimCod,CodeOpenAI,MathAI}. The \textit{measuring massive multitask language understanding (MMLU)} approach \cite{mmlu} evaluates LLMs across a wide range of subjects, including mathematics, history, and computer science. The initial results in \cite{mmlu} showed that the evaluated models were particularly weak in socially important areas such as morality and law. More recent studies, including MMLU-Pro \cite{mmlu_pro} and MMLU-Redux \cite{mmlu_redux}, demonstrate substantial improvements in LLM performance while also highlighting remaining limitations and considerable variation across domains. MMLU-Pro further shows that more challenging, reasoning-intensive questions remain difficult for modern LLMs, while MMLU-Redux identifies errors in the original MMLU benchmark that can affect reported performance. The \textit{beyond the imitation game benchmark (BIG-bench)} evaluates language models across 204 diverse tasks designed to be challenging for existing models. The authors investigate whether performance trends observed in smaller models can be used to predict the capabilities of larger models \cite{bigbench}. Their results show that model performance generally improves with scale, but the rate and pattern of improvement vary substantially across tasks. The study also shows that apparent capability breakthroughs can depend on how performance is measured. This highlights the importance of evaluation metrics and separating distinct failure modes when assessing LLMs.

In addition to LLMs performance evaluation on generic subjects, several recent studies have focused on evaluation of LLMs as mathematical problem solvers. Cobbe et al. introduce the GSM8K dataset and evaluate the mathematical capability of LLMs when tested on this dataset. The results show that LLMs can perform multi-step mathematical reasoning. Interestingly, the verification mechanism they proposed can substantially improve answer accuracy \cite{gsm8k}. This can follow from the fact that many inconsistencies in the LLM responses are due to \textit{execution errors}. The study, however, does not evaluate the scalability of this behaviour, i.e., to \reviewchange{characterize} the performance scaling with problem dimension. Recent studies in \cite{lin2025zebralogic,GSM} elaborate the importance of such evaluations, as \reviewchange{they} provide means to distinguish different failure modes.

\section{Proposed Framework}

\label{sec:framework}
We present our framework by first describing the proposed LLM-assisted optimization protocol with iterative refinement loop. The proposed protocol enables us to distinguish between execution error and failure mode of LLMs. We then present the representative optimization families and deterministic reference solvers and define the key evaluation metrics.

\subsection{LLM-assisted Optimization with Iterative Refinement}
\label{subsec:llm_protocol}
Consider a generic constrained optimization of form
\begin{subequations}
\begin{align}
    &\min_{\mathbf{x}} f(\mathbf{x}) 
    \qquad \text{s.t. } &&g_i(\mathbf{x}) \leq 0 \text{ for } i\in \{ 1,\ldots, I\}, \\
    & &&h_j(\mathbf{x}) = 0
    \text{ for } j\in \{ 1,\ldots, J\}.
\end{align}
\end{subequations}
We denote the problem dimension, i.e., the dimension of $\mathbf{x}$, by $K$. The proposed framework deploys LLMs to estimate a solution to this problem by first prompting an LLM to generate candidate solutions, and then utilizing an LLM to iteratively refine the solution with the help of an external evaluator. We go through the details in the sequel.

\subsubsection{Prompt Design}
For the initial prompt, we provide the following information to the LLM:
\begin{enumerate}[label=\alph*.]
    \item general description of problem,
    \item problem structure along with key corresponding properties, e.g., \textit{``concavity implies more balanced levels yield higher rates''} for sum-rate maximization,
    \item an example to illustrate the calculation and output format,
    \item detailed problem parameters, e.g., problem dimension $K$, exact shape of the objective and constraints,
    \item return format specification, e.g., a JSON list of $K$ nonnegative numbers.
\end{enumerate}
A full example of the initial prompt for sum-rate optimization is given in \promptlocation{}.

\subsubsection{Output Formatting Requirements}
LLMs are instructed to return a list of \textit{exactly} $K$ variables. i.e., $\mathbf{x} = [x_1, x_2, \ldots, x_K]$. We parse the response and check its format. A \textit{valid output} must be a JSON list with $K$ variables.

\subsubsection{Parsing and Feasibility Check}
A \textit{parsing mechanism} is implemented to extract the output from the response and confirm its validity. Once the validity of the output is confirmed, we check its feasibility using the problem constraints.

\subsubsection{Iterative Refinement}
\label{subsec:iterative_refinement}
Once the one-shot response of the LLM is confirmed, we  apply an iterative refinement protocol to detect failure or suboptimal solutions and provide feedback. The feedback contains two informative components: the \textit{failure mode}, which informs the LLM about possible failure in the candidate solution, and the \textit{evaluation feedback}, which evaluates a feasible and valid candidate solution. Possible failure modes are
%After receiving the first LLM output, we check whether it is a valid and feasible solution. The following failure modes are considered:
\begin{enumerate}[label=\alph*.] 
    \item \textit{Format Error.} This mode indicates that output does not satisfy the formatting requirements. %return %a response as required in the prompt.
    \item \textit{Constraint Violation.} It indicates that output does not meet the constraints. Given potential execution errors, we activate this mode for each type of constraint differently. 
    \begin{enumerate}
      \item[($i$)] For inequality constraints, this mode is activated if the constraint is violated with a tolerable margin $\varepsilon$, i.e., if
      \begin{align}
          \max_{i \in \{1, \ldots,I\} }  \reviewchange{g_i(\mathbf{x})} > \varepsilon.
      \end{align}
      A typical choice is $\varepsilon = 10^{-6}$. %sum-rate and energy-efficiency problems, if the total power exceeds the budget, we consider this as an \emph{overbudget error}.
      \item[($ii$)] For equality constraints, this mode is activated if 
      \begin{align}
        \max_{j \in \{1, \ldots,J\} } \vert h_j(\mathbf{x})\vert > \tau,    
      \end{align}
      for some tolerable margin $\tau$. A typical interval for $\tau$ is $\tau \in [10^{-4}, 10^{-2}]$.
      %the projection problem, the equality constraint $A\mathbf{x} + \mathbf{b} = \mathbf{0}$ is checked against a configurable tolerance $\tau$: a solution is considered valid if $\max_j |[A\mathbf{x} + \mathbf{b}]_j| \leq \tau$. We vary $\tau \in \{10^{-4}, 10^{-3}, 10^{-2}\}$ to study how constraint precision affects LLM performance. The inequality constraint $U\mathbf{x} \leq \mathbf{d}$ and nonnegativity constraint are checked with a fixed tolerance of $10^{-6}$.
    \end{enumerate}
    \item \textit{Other Errors.} Errors such as network issues, interrupted API calls, or unexpected system-level failures that are not related to the LLM response are included in this mode.
    \item \textit{No Error.} This mode confirms the validity and feasibility of the solution.
\end{enumerate}
If a failure is detected, the loop provides feedback and requests a new candidate solution. 

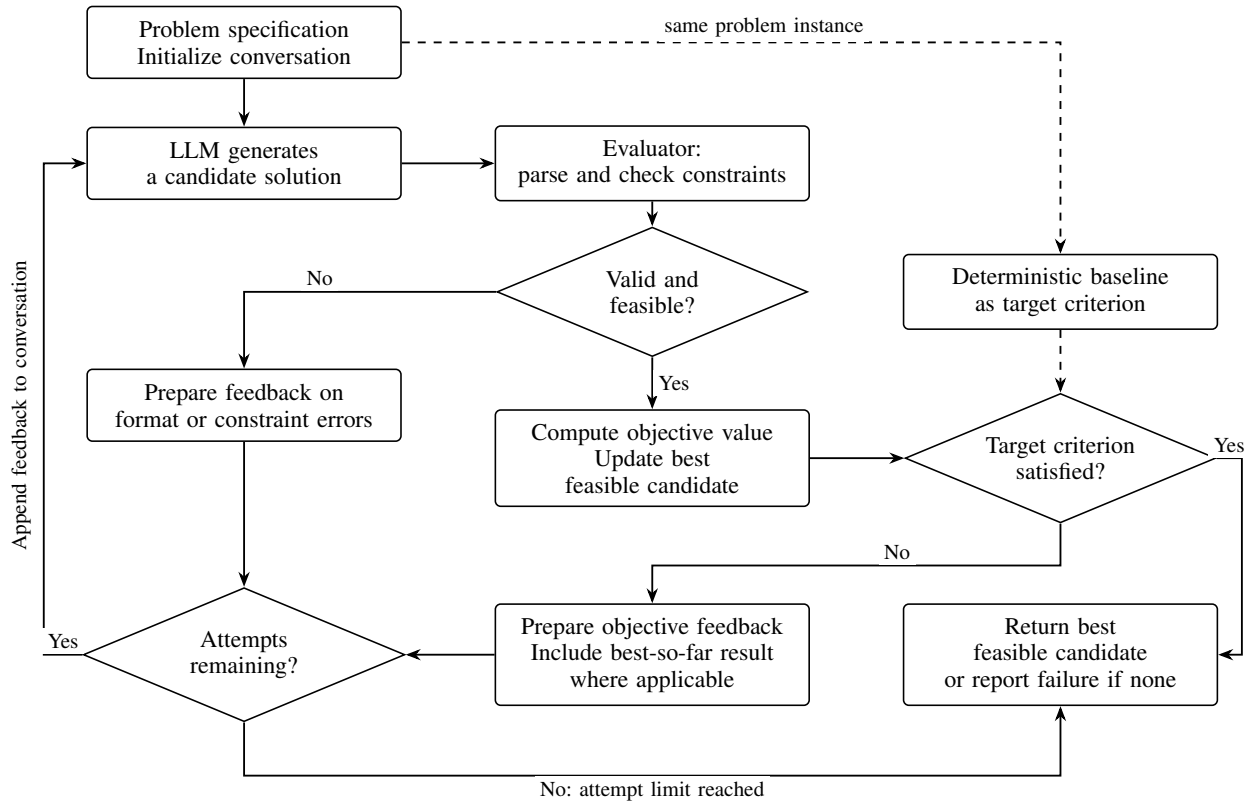
\begin{figure*}[t]
\centering
% Shared refinement flow; task-specific stopping tolerances remain in the code.
% The deterministic reference is used only by the external evaluator.
\begingroup
\colorlet{frameworkInk}{black}
% \ifshowrevisions
%   \colorlet{frameworkInk}{revisionGreen}
% \else
%   \colorlet{frameworkInk}{black}
% \fi
\begin{tikzpicture}[
  font=\small, text=frameworkInk,
  box/.style={draw=frameworkInk, rounded corners=2pt, line width=0.6pt,
    align=center, text width=3.8cm, minimum height=0.95cm, inner sep=5pt},
  decision/.style={draw=frameworkInk, diamond, aspect=2.4, line width=0.6pt,
    align=center, text width=2.5cm, inner sep=1pt},
  flow/.style={-{Stealth[length=2mm]}, draw=frameworkInk, line width=0.7pt},
  reference/.style={flow, dashed},
  lab/.style={font=\footnotesize, fill=white, inner sep=2pt}
]
\node[box] (prompt) at (0,0) {Problem specification\\Initialize conversation};
\node[box] (llm) at (0,-1.6) {LLM generates\\a candidate solution};
\node[box] (validate) at (5.4,-1.6) {Evaluator:\\parse and check constraints};
\node[decision] (valid) at (5.4,-3.3) {Valid and\\feasible?};
\node[box] (failure) at (0,-4.8) {Prepare feedback on\\format or constraint errors};
\node[box] (score) at (5.4,-5.5) {Compute objective value\\Update best feasible candidate};
\node[box] (baseline) at (10.8,-3.3) {Deterministic baseline as target criterion};
\node[decision] (target) at (10.8,-5.5) {Target criterion\\satisfied?};
\node[box] (feedback) at (5.4,-8.1) {Prepare objective feedback\\Include best-so-far result\\where applicable};
\node[decision] (budget) at (0,-8.1) {Attempts\\remaining?};
\node[box] (output) at (10.8,-8.1) {Return best feasible candidate\\or report failure if none};
\draw[flow] (prompt) -- (llm);
\draw[reference] (prompt.east) -- node[lab,above,pos=0.55] {same problem instance} (10.8,0) -- (baseline.north);
\draw[flow] (llm) -- (validate);
\draw[flow] (validate) -- (valid);
\draw[flow] (valid.west) -| node[lab,above,pos=0.35] {No} (failure.north);
\draw[flow] (valid) -- node[lab,right] {Yes} (score);
\draw[flow] (score) -- (target);
\draw[reference] (baseline) -- (target);
\draw[flow] (target.east) -- node[lab,above] {Yes} (13.2,-5.5) |- (output.east);
% Keep the return branch above the feedback box even when its text reflows.
\coordinate (feedback-entry) at ([yshift=5mm]feedback.north);
\draw[flow] (target.south) |- node[lab,above,pos=0.7] {No} (feedback-entry) -- (feedback.north);
\draw[flow] (feedback) -- (budget);
\draw[flow] (failure) -- (budget);
\draw[flow] (budget.west) -- node[lab,above] {Yes} (-2.65,-8.1) -- (-2.65,-1.6) -- (llm.west);
\node[font=\footnotesize,align=center,rotate=90] at (-2.95,-4.8) {Append feedback to conversation};
\draw[flow] (budget.south) -- (0,-9.7) -- node[lab,below] {No: attempt limit reached} (10.8,-9.7) -- (output.south);
\end{tikzpicture}
\endgroup
\caption{Schematic of the LLM-assisted optimization framework. Validation and objective evaluation are performed by a deterministic evaluator. Dashed arrows show the deterministic baseline path, which supplies only the external stopping target. Refinement ends when the target criterion is satisfied or the attempt limit is reached.}
\label{fig:refinement-framework}
\end{figure*}

When no failure is detected, the feasible solution is evaluated using its objective value. The comparison can be \textit{baseline-aided,} where we compare the result to a deterministic optimum computed by a baseline algorithm. The baseline result serves as an external target and stopping criterion and is not shown to the LLM. If the optimality criterion is not satisfied, the candidate's objective value and, where applicable, the best objective value among previously generated feasible candidates are provided as feedback to the LLM through textual prompts. Table~\ref{tab:prompts} gives some feedback prompt examples. The iterative refinement is executed in a conversational format. Each new feedback prompt is appended to a conversation containing previous prompts and proposed solutions. This ensures that the LLM has sufficient context to refine its output and correct the execution errors. A schematic of the proposed framework is given in Figure~\ref{fig:refinement-framework}.

\begin{table}[t]
\caption{Examples of feedback prompts used in the refinement loop of the sum-rate maximization problem.}
\label{tab:prompts}
% \ifshowrevisions\color{revisionGreen}\fi
\centering
\renewcommand{\arraystretch}{1.3}
\begin{tabularx}{\linewidth}{l >{\raggedright\arraybackslash}X}
\hline
\textbf{invalid format} & \ttfamily Your output is not a valid JSON list. Return a JSON list of K nonnegative numbers summing to P. Try again. \\
\hline
\textbf{invalid result length} & \ttfamily Your output has <\ldots> items, but K=<\ldots>. Return a list of exactly K nonnegative numbers summing to P. Try again. \\
\hline
\textbf{overbudget error} & \ttfamily With solution <\ldots>, total power=<\ldots> and is over budget. The total power must be less than <\ldots>. Try again. \\
\hline
\textbf{suboptimal solution} & \ttfamily With solution <\ldots> you get <\ldots>. Best assignment so far produces <\ldots>. Try adjusting solution to balance levels to achieve better result, but do not repeat solutions you have already seen. \\
\hline
\textbf{general feedback} & \ttfamily With solution <\ldots> you get <\ldots>. Try adjusting solution to balance levels to achieve better result, but do not repeat solutions you have already seen. \\
\hline
\end{tabularx}
\end{table}

\subsection{Representative Optimization Families}
We evaluate our framework on three optimization families, namely known convex optimization, fractional programming, and general-form constrained optimization. For each family, we consider a representative problem and a baseline solver. The problems and their solvers are given in the sequel. 
\subsubsection{Convex Optimization}
For this family, we consider the problem of sum-rate maximization in wireless communication networks \cite{el2011network,waterfilling1} with $K$ parallel Gaussian channels: assume link $i$ has real-valued gain $g_i>0$. Assuming additive zero-mean Gaussian noise in the channel with variance $N_i>0$, the quality of this link can be assessed by its \textit{capacity} as \cite{shannon}
\begin{align}
    R_i (x_i) = \log_2 \left( 1+ \frac{x_i}{f_i} \right),
\end{align}
where $f_i$ denotes the inverse of signal-to-noise ratio (SNR) gain in link $i$, defines as
\begin{equation}
f_i \triangleq \frac{N_i}{g_i},
\end{equation}
and $x_i \ge 0$ denotes the transmit power allocated to link $i$. The power is allocated subject to a total budget, i.e., 
\begin{equation}
\sum_{i=1}^{K} x_i \leq P_{\text{total}},
\end{equation}
for some total power budget $P_{\text{total}} > 0$ \cite{el2011network}.

% For each channel, the resulting signal-to-noise ratio (SNR) is
% \begin{equation}
% \gamma_i \triangleq \frac{g_i x_i}{N_i} = \frac{x_i}{f_i},
% \end{equation}
% and we define $L_i \triangleq 1+\gamma_i = 1+\frac{x_i}{f_i}$.
% Under the Shannon--Hartley theorem, the achievable rate for a channel is proportional to $\log_2(1+\gamma_i)$, which motivates the sum-rate objective used in this work .

To optimize the system throughput, we maximize the \textit{sum-rate} over the $K$ parallel links subject to the power budget. We can formulate this problem as
\begin{subequations}
\begin{align}
\label{eq:waterfilling}
& \max_{\mathbf{x}} \quad
R(\mathbf{x})
= \sum_{i=1}^{K} \log_2\!\left(1 + \frac{x_i}{f_i}\right),\\
&\text{s.t.} \; \sum_{i=1}^{K} x_i \leq P_{\text{total}}, \; x_i \geq 0, \quad \text{ for } i\in\{1, \ldots,K\}.
\end{align}
\end{subequations}
% subject to
% \begin{equation}
% .
% \end{equation}
% where $f_i = N_i/g_i$ is the noise-to-gain ratio. The logarithmic form follows from the Shannon capacity of a channel.
The objective $R(\mathbf{x})$ in \eqref{eq:waterfilling} is concave because it is a sum of concave terms. It therefore has diminishing marginal returns. Allocating additional power to a channel with a larger SNR produces a smaller rate increase than allocating the same power to a channel with a smaller SNR. %This concavity encourages balancing across channels when power is sufficient. Sum-rate maximization over parallel Gaussian channels under a total power constraint 
The optimization in \eqref{eq:waterfilling} is a classical problem in wireless communications. For this problem, the Karush-Kuhn-Tucker (KKT) conditions yield the well-known \emph{water-filling} solution \cite[p.~245]{boyd2004convex}, where the optimal transmit power of link $i$ is given by
\begin{equation}
    \label{eq:waterfilling_solution}
x_i^\star = \max \left\lbrace\mu - \frac{N_i}{g_i}, 0\right\rbrace,
\end{equation}
for the water level $\mu$ that is chosen such that 
\begin{align}
\sum_i x_i^\star = P_{\text{total}}.    
\end{align}
This solution provides an efficient baseline.

The sum-rate maximization problem tests whether an LLM can infer and deploy this well-known structure in the  resource-allocation problem. The problem is also scalable in $K$, allowing us to study how LLM performs with the problem size.

\subsubsection{Fractional Programming}
Fractional programming is a class of generally non-convex problems. Although non-convex, there are various approaches to efficiently approximate their solutions \cite{dinkelbach1967,schaible1976fractional,fp_part1,fp_part2}. For this family, we consider the problem of energy-efficiency maximization: considering the sum-rate $R(\mathbf{x})$ in \eqref{eq:waterfilling}, we can define the \textit{energy-efficiency} as the sum-rate per unit of consumed power in the system, i.e., 
\begin{equation}
\eta(\mathbf{x})
= \frac{R(\mathbf{x})}{
\displaystyle 
P_{\text{const}} + \sum_{i=1}^{K} x_i}, 
\end{equation}
where $P_{\text{const}} \ge 0$ denotes the fixed power dissipation in the system, e.g., circuit power consumption. Considering energy-efficiency as the throughput, we can formulate the throughout optimization problem as
\begin{subequations}
\begin{align}
\label{eq:efficiency}
&\max_{\mathbf{x}} \quad
%\eta(\mathbf{x})
%= \frac{R(\mathbf{x})}{P_{\text{const}} + \sum_{i=1}^{K} x_i}
%= 
\frac{
\displaystyle R(\mathbf{x})
}
{
\displaystyle 
P_{\text{const}} + \sum_{i=1}^{K} x_i
},\\ %\sum_{i=1}^{K} \log_2\!\left(1 + \frac{x_i}{f_i}\right),\\
    \label{eq:power_constraint}
&\text{s.t.} \; \sum_{i=1}^{K} x_i \leq P_{\text{total}}, \; x_i \geq 0, \quad \text{ for } i\in\{1, \ldots,K\}.
\end{align}
\end{subequations}
% subject to the same constraints in \eqref{eq:power_constraint}.
% Here, $P_{\text{const}} \ge 0$ denotes a constant static power dissipation (e.g., circuit power consumption), so that the denominator represents the total power consumption.

In \eqref{eq:efficiency}, $R(\mathbf{x})$ is concave; however, the fractional objective $\eta(\mathbf{x})$ is generally not concave. The problem belongs to the standard class of \textit{concave fractional programs} and can be solved via Dinkelbach's method \cite{dinkelbach1967,fp_part1}. The key idea is to iteratively parameterize the ratio by a scalar $\lambda_t$ and solve the parameterized subproblem
\begin{equation}
\label{eq:dinkelbach_subproblem}
\max_{\mathbf{x}} \quad
\Bigl[R(\mathbf{x}) - \lambda_t \bigl(P_{\text{const}} + \sum_{i=1}^{K} x_i\bigr)\Bigr]
\quad \text{s.t. } \eqref{eq:power_constraint}.
\end{equation}
The baseline solver for this problem is illustrated in the next subsection. 
% For the separable logarithmic objective in our setting, the KKT conditions of
% \eqref{eq:dinkelbach_subproblem} imply a water-filling structure, i.e., for each iteration $t$
% \begin{equation}
% x_i^\star = \max \{ \mu_t - f_i, 0 \},
% \end{equation}
% for some water level $\mu_t$. If the total-power constraint is inactive, the stationarity gives the
% \emph{unconstrained} water level
% \begin{equation}
% \mu_t^{\mathrm{uncon}} = \frac{1}{\lambda_t \ln 2}.
% \end{equation}
% %and $x_i^\star = \max (\mu_k^{\mathrm{uncon}} - f_i, 0)$.
% If the total-power constraint is active, an additional dual variable
% $\nu_t \ge 0$ appears in the stationarity condition, yielding
% \begin{equation}
% \mu_t = \frac{1}{(\lambda_t + \nu_t)\ln 2},
% \end{equation}
% where $\mu_t$, and hence $\nu_t$, is chosen such that the power budget is met, i.e., 
% \begin{equation}
% \sum_{i=1}^K \max\{\mu_t - f_i, 0\} = P_{\text{total}}.
% \end{equation}

% After solving for $\mathbf{x}_k$, we update $\lambda_{t+1}$ as the ratio of the current objective value to the current denominator, i.e., 
% \begin{equation}
% \lambda_{t+1} = \frac{R(\mathbf{x}_k)}{
% \displaystyle
% P_{\text{const}} + \sum_{i=1}^K x_i^{(t)}}.
% \end{equation}
% The iteration converges to the optimal energy efficiency when the optimality condition is satisfied.

Energy efficiency optimization is more complex than sum-rate maximization, since it deals with a ratio objective. The classical solution also builds a sequence of converging water-filling problems. We use this problem to test whether the LLM can recognize this structure, and whether it deploys the connections between this task and the water-filling problem. %can refine its this recognition through feedback about the current objective. %while still satisfying the same convex constraints.

\subsubsection{General-Form Optimization}
As the third family, we look into general-form optimizations, which unlike the previous two classes do not have a well-known standard solver. For this class, we consider the problem of projecting a target point onto a convex feasible set. In addition to the inequality constraints, this problem includes equality constraints, which did not exist in the former problems. 

Consider a randomly generated target point $\mathbf{y} \in \mathbb{R}^K$. Let $(A, \mathbf{b}) \in \mathbb{R}^{p \times K} \times \mathbb{R}^p$ and $(U, \mathbf{d})\in \mathbb{R}^{m \times K} \times \mathbb{R}^m$ specify two affine transforms that map $\mathbf{x}\in \mathbb{R}^K$ to points in $\mathbb{R}^p$ and \reviewchange{$\mathbb{R}^m$}, respectively. The optimization problem is given by
\begin{subequations}
\begin{align}
\label{eq:projection_formulation}
&\min_{\mathbf{x}} \quad \Vert \mathbf{x}-\mathbf{y}\Vert^2\\ %\sum_{i=1}^{K} (x_i - y_i)^2\\
&\text{s.t. }\;
x_i \geq 0, \quad \text{ for } i \in \{1,\ldots,K\}, \label{eq:proj_nonneg}\\
&\hphantom{\text{s.t. }\;} U\mathbf{x} - \mathbf{d} \leq \mathbf{0}, \label{eq:proj_inequality}\\
&\hphantom{\text{s.t. }\;} A\mathbf{x} + \mathbf{b} = \mathbf{0}, \label{eq:proj_equality}
\end{align}
\end{subequations}

To instantiate this problem, we first sample a strictly feasible point, and then %$\mathbf{x}_{\text{feas}} > 0$, then 
construct $(A, \mathbf{b})$ and $(U,\mathbf{d})$, such that %$\mathbf{x}_{\text{feas}}$ 
the feasible point satisfies all constraints. This guarantees that the feasible set is nonempty, and hence the problem is well-posed.

The problem in \eqref{eq:projection_formulation} is a general convex quadratic program with mixed constraints that can be solved deterministically by a convex solver. Note that former representative problems have well-known solvers that appear frequently in wireless communications literature \cite{boyd2004convex,fp_part1,dinkelbach1967}. Unlike them, the projection problem %in  is . While projection problems are well-studied in convex optimization theory, the 
with specific random instances does not correspond to any standard textbook formulation that LLMs would likely have memorized. As a result, this problem provides a reference to evaluate the level of understanding the LLMs can obtain from the problem structure. 

A key consideration in our framework is the tolerance $\tau$ used to check the equality constraint in \eqref{eq:proj_equality}. %$A\mathbf{x} + \mathbf{b} = \mathbf{0}$. 
Since LLMs produce outputs with limited numerical precision, requiring exact equality can lead to dominance of execution error in evaluation. We therefore accept a solution as satisfying the equality constraint if $\max_j |[A\mathbf{x} + \mathbf{b}]_j| \leq \tau$. We further study the effect of $\tau$ on validity and optimality rates and tune it for minimal impact of execution errors in the evaluation.

\subsection{Reference Solvers}
For the first two problems, the baseline solver is given by directly implementing the given solution. For the latter one, one can deploy various approaches. In the sequel, we briefly elaborate the baseline solver for each problem. 
\subsubsection{Water-Filling Solver}
For sum-rate maximization, we directly implement the water-filling algorithm. % that solves the problem analytically. 
The algorithm sorts the noise floors $f_i$ in ascending order and computes their \textit{prefix sums}. We determine the number of active channels $m^\star$ by testing candidate values $m \in \{1, \ldots, K\}$. For each candidate $m$, we compute the corresponding water level as
\begin{equation}
\mu_m = \frac{P_{\text{total}}}{m}
 + \frac{1}{m} \sum_{j=1}^m f_{(j)},
\end{equation}
where $f_{(j)}$ denotes the $j$-th smallest floor. Using the resulting water level $\mu_m$, we check if the assigned powers to the $m$ active channels are positive. The unique consistent value of $m$ is used to compute the solution as $x_i^* = \max\{0, \mu^* - f_i\}$. The algorithm runs in $O(K \log K)$ time, dominated by sorting.

\subsubsection{Fractional Solver}
%Fractional programming (FP) is a general framework for optimization problems with ratio-based objectives. \emph{Fractional Programming for Communication Systems---Part I: Power Control and Beamforming}\cite{fp_part1} provides a systematic overview of FP methods for communication systems. It covers classical single-ratio problems as well as more general ratio structures that arise in power control and beamforming.
We implement Dinkelbach's method to solve the energy-efficiency problem. The algorithm initializes a scalar \reviewchange{$\lambda_0$} from an equal-power solution, and iteratively solves the parameterized subproblem in \eqref{eq:dinkelbach_subproblem}.

For the separable logarithmic objective in our setting, the  Karush–Kuhn–Tucker (KKT) conditions of
\eqref{eq:dinkelbach_subproblem} imply a water-filling structure, i.e., for each iteration $t$ \cite{boyd2004convex}
\begin{equation}
\reviewchange{x_i^{(t)}} = \max \{ \mu_t - f_i, 0 \},
\end{equation}
for some water level $\mu_t$. If the power constraint is inactive, the stationarity gives the
\emph{unconstrained} water level
\begin{equation}
\mu_t^{\mathrm{uncon}} = \frac{1}{\lambda_t \ln 2}.
\end{equation}
%and $x_i^\star = \max (\mu_k^{\mathrm{uncon}} - f_i, 0)$.
If the power constraint is active, an additional dual variable
$\nu_t \ge 0$ appears in the stationarity condition, yielding
\begin{equation}
\mu_t = \frac{1}{(\lambda_t + \nu_t)\ln 2},
\end{equation}
where $\mu_t$, and hence $\nu_t$, is chosen such that the power budget is met, i.e., 
\begin{equation}
\sum_{i=1}^K \max\{\mu_t - f_i, 0\} = P_{\text{total}}.
\end{equation}
After solving for \reviewchange{$\mathbf{x}^{(t)}$}, we update $\lambda_{t+1}$ as the ratio of the current objective value to the current denominator, i.e.,
\begin{equation}
\lambda_{t+1} = \frac{\reviewchange{R(\mathbf{x}^{(t)})}}{
\displaystyle
P_{\text{const}} + \sum_{i=1}^K x_i^{(t)}}.
\end{equation}
The iteration converges to the optimal energy efficiency when the optimality condition is satisfied.
% \begin{equation}
% \begin{split}
% \max_{\mathbf{x}} \Bigl[R(\mathbf{x}) - \lambda_k (P_{\text{const}} + \sum_i x_i)\Bigr] \\
% \text{s.t.} \quad \sum_i x_i \le P_{\text{total}}, \, x_i \ge 0.
% \end{split}
% \end{equation}
% For the separable logarithmic rate function, the KKT conditions of this subproblem yield a water-filling solution. Specifically, the stationarity condition yields the water level $\mu = 1/(\lambda_k \ln 2)$, and the optimal allocation takes the form $x_i = \max(0, \mu - f_i)$. 
% If the unconstrained solution $\mathbf{x}_{\text{uncon}}$ violates the power budget ($\sum_i x_i^{\text{uncon}} > P_{\text{total}}$), we perform bisection on $\mu$ to find the largest value satisfying $\sum_i \max(0, \mu - f_i) \le P_{\text{total}}$.
% After obtaining the solution $\mathbf{x}_k$ for iteration $k$, we compute the objective and denominator values:
% \begin{equation}
% f_k = \sum_i \log_2(1 + x_i^{(k)}/f_i), \quad g_k = \sum_i x_i^{(k)} + P_{\text{const}},
% \end{equation}
% % and update $\lambda_{k+1} = f_k / g_k$. 
% In our implementation, the iteration terminates when the optimality condition
% \begin{equation}
%     |f_k - \lambda_k g_k| < \epsilon = 10^{-9}
% \end{equation}
% is satisfied. 

Convergence of the algorithm is guaranteed by the properties of Dinkelbach's method \cite{dinkelbach1967}. Throughout our experiments, the algorithm typically converges within 20 to 50 iterations. Our implementation caps the number of iterations at 200.

\subsubsection{ADMM Solver}
For the convex projection problem, we use the \textit{alternating direction method of multipliers (ADMM)} as the baseline solver. ADMM is a general-purpose algorithm for solving convex optimization problems with separable structure and linear constraints \cite{admm}. By decomposing a constrained optimization into smaller sub-problems, ADMM efficiently solves mixed constrained problems, i.e., problems with both equality and inequality constraints.
% In our work, we use ADMM as the deterministic baseline for the minimum-distance projection problem, where the mixed constraint structure makes it a natural algorithmic choice.

For the optimization in \eqref{eq:projection_formulation}, ADMM reformulates the problem by introducing auxiliary variables \revision{$\mathbf{z}_1 \in \mathbb{R}^K$ and $\mathbf{z}_2 \in \mathbb{R}^m$, with $\mathbf{x}=\mathbf{z}_1\geq\mathbf{0}$ and $U\mathbf{x}-\mathbf{d}=\mathbf{z}_2\leq\mathbf{0}$, corresponding to the nonnegativity constraint in \eqref{eq:proj_nonneg} and inequality constraint in \eqref{eq:proj_inequality}, respectively. The equality constraint $A\mathbf{x}+\mathbf{b}=\mathbf{0}$ is enforced directly in the $\mathbf{x}$-update.} The augmented Lagrangian with scaled dual variables \revision{$\mathbf{u}_1\in\mathbb{R}^K$, $\mathbf{u}_2\in\mathbb{R}^m$} and penalty parameter $\rho$ leads to the following iterative updates.

\textbf{$\mathbf{x}$-update.} Solve the linear system of equations
\begin{subequations}
\begin{align}
M\mathbf{x}^{t+1} ={}& 2\mathbf{y}+\rho(\mathbf{z}_1^t-\mathbf{u}_1^t)
\\
&+\rho U^\top(\mathbf{z}_2^t-\mathbf{u}_2^t+\mathbf{d})-A^\top\boldsymbol{\nu}^{t+1},
\nonumber
\\
A\mathbf{x}^{t+1} &= -\mathbf{b}.
\end{align}
\end{subequations}
for $\mathbf{x}^{t+1}$ \revision{and $\boldsymbol{\nu}^{t+1}$}, where \revision{$M=(2+\rho)I+\rho U^\top U$}. \revision{The $\mathbf{x}$-update introduces the Lagrange multiplier $\boldsymbol{\nu}^{t+1}$ to minimize the augmented Lagrangian subject to the equality constraint $A\mathbf{x}+\mathbf{b}=\mathbf{0}$.} We pre-compute a Cholesky factorization of $M$ for fast computation across iterations.
\revision{At each iteration, $\boldsymbol{\nu}^{t+1}$ is determined jointly with $\mathbf{x}^{t+1}$ by these two equations and requires no separate dual update.}

\textbf{$\mathbf{z}$-updates.} Each auxiliary variable is updated via a proximal operator that enforces its corresponding constraint as
\begin{subequations}
    \begin{align}
\mathbf{z}_1^{t+1} &= \max\{\mathbf{x}^{t+1} + \revision{\mathbf{u}_1^t},\, \mathbf{0}\}, \\
\revision{\mathbf{z}_2^{t+1}} &\revision{= \min\{U\mathbf{x}^{t+1} - \mathbf{d} + \mathbf{u}_2^t,\, \mathbf{0}\}.}
\end{align}
\end{subequations}

\textbf{$\mathbf{u}$-updates.} The scaled dual variables accumulate constraint residuals as
\begin{subequations}
\begin{align}
\mathbf{u}_1^{t+1} &= \mathbf{u}_1^t + \mathbf{x}^{t+1} - \mathbf{z}_1^{t+1}, \\
\revision{\mathbf{u}_2^{t+1}} &\revision{= \mathbf{u}_2^t + U\mathbf{x}^{t+1} - \mathbf{d} - \mathbf{z}_2^{t+1}.}
\end{align}
\end{subequations}
The algorithm terminates when both the primal residual and the iterate change fall below $10^{-5}$.

\subsection{Evaluation Metrics}
\label{subsec:eval_metrics}
%For each problem family, 
We evaluate the performance of LLM-based solver using the a set of metrics described below. These metrics enable direct analysis of scaling with the problem dimension $K$ and impact of different failure modes. % affect achievable objective values.

\subsubsection{Validity Rate}
If the LLM encounters one of the failure modes mentioned in Section~\ref{subsec:iterative_refinement}, we log it as \textit{invalid}. The \textit{validity rate} is then defined as the fraction of valid outputs over the total number of attempts.
\begin{equation}
    \mathrm{VR} = \frac{\text{\# valid outputs}}{\text{\# total attempts}}.
\end{equation}

\subsubsection{Optimality Rate}
The \textit{optimality rate} is defined as the fraction of valid outputs that are \textit{optimal},\footnote{For the fractional programming problem, we mis-use the term \textit{optimal} to refer to the solution given by Dinkelbach's method.} i.e., they achieve the deterministic solution given by the reference solver. To compare the two solutions, we consider an error tolerance of $\tau$, often $\tau = 10^{-4}$. We can hence write
\begin{equation}
    \mathrm{OR} = \frac{\text{\# optimal outputs}}{\text{\# valid outputs}}.
\end{equation}

\subsubsection{End-to-end Optimality Rate}
The \textit{end-to-end optimality rate} is defined as the fraction of all attempts that produce \textit{valid and optimal} outputs, i.e.,
\begin{equation}
    \mathrm{OR}_{\mathrm{E2E}} = \frac{\text{\# optimal outputs}}{\text{\# total attempts}}.
\end{equation}
It is easy to see that
\begin{equation}
    \mathrm{OR}_{\mathrm{E2E}} = \mathrm{VR} \cdot 
    \mathrm{OR}.
\end{equation}
\subsubsection{Failure Mode Frequency}
To analyze how failure patterns scale with problem dimension $K$, we categorize the unsuccessful attempts based on the specific failure mode by evaluating the frequency of each mode.

\subsubsection{Performance Ratio}
The \textit{performance ratio} measures how close the best LLM solution is to the optimal solution.
For valid outputs, we define this metric as the ratio of the objective value computed by the LLM solver to the one given by the deterministic solver. More precisely, for the rate-maximization task, the performance rate is given by
\begin{equation}
    \mathrm{PR} =
    \frac{R(\hat{\mathbf{x}})}{R(\mathbf{x}^\star)},
\end{equation}
for the energy-efficiency maximization it reads
\begin{equation}
    \mathrm{PR} =
    \frac{\eta(\hat{\mathbf{x}})}{\eta(\mathbf{x}^\star)},
\end{equation}
and for the projection task, it is given by\footnote{Note that this is a minimization problem.}
\begin{equation}
    \mathrm{PR} =
    \frac{\Vert \mathbf{x}^\star- \mathbf{y}\Vert^2}{\Vert\hat{\mathbf{x}}- \mathbf{y}\Vert^2},
\end{equation}
where $\hat{\mathbf{x}}$ is the best valid LLM-based solution, and $\mathbf{x}^\star$ denotes the reference solution. %,and $D$ are the respective objective functions. 
If the ratio is less than 1 given a tolerance of $\varepsilon= 10^{-4}$, i.e. $\mathrm{PR} < 1 - \varepsilon$, we consider the LLM solution to be suboptimal.

\subsubsection{Runtime}
% For sum-rate and energy-efficiency problems, 
In addition to performance metrics, we report the runtime of the LLM-based optimization process.

Classical deterministic solvers are computationally efficient. For $K$ parallel channels, water-filling can be implemented in $O(K\log K)$ time, or in $O(K)$ time with specialized selection routines. Dinkelbach's method typically requires a moderate number of iterations. Each iteration solves a convex subproblem with a comparable computational cost. In practice, these baselines often run in the order of milliseconds. In contrast, LLM-based optimization incurs substantially higher latency per iteration, often on the order of seconds. Execution time has three main components: LLM generation, API communication, and local code execution. Local execution includes tasks such as scenario setup and solution recording. The latter overhead is small compared with model generation and API communication, so we treat it as negligible.

\section{Experimental Investigation}
We now investigate the proposed framework and characterize the behaviour of the LLM-based solvers. 
\label{sec:experiments}
\subsection{Experimental Setup}
\subsubsection{Evaluated LLMs}
\revision{We use two LLMs in our experiments: OpenAI's o4-mini~\cite{o4mini_model} and Google's Gemini 3 Flash Preview (\texttt{gemini-3-flash-preview})~\cite{gemini3flash_model}.} Hereafter, we denote the models by \omini{} and \gem{}, respectively. Both models are accessed via their respective APIs with the settings described in Table~\ref{tab:llm-setup}. For this experiment, we do not consider methods to mitigate the limitation of context window size (e.g., context window management).

\begin{table}[t]
\caption{Context size and key settings in the experiments.}
\centering
\renewcommand{\arraystretch}{1.3}
\setlength{\tabcolsep}{4pt}
\begin{tabularx}{\linewidth}{@{} l >{\raggedright\arraybackslash}X r @{}}
\toprule
\textbf{Model} & \textbf{Parameter} & \textbf{Setting} \\
\midrule
o4-mini & Max window size & 100,000 \\
\midrule
gemini-3-flash-preview & Max window size ($K\le 160$) & 100,000 \\
& Max window size ($160 < K \le 400$) & 200,000 \\
& Max window size ($K > 400$) & 400,000 \\
& Thinking level & \texttt{HIGH} \\
\bottomrule
\end{tabularx}
\label{tab:llm-setup}
\end{table}

The sum-rate and energy-efficiency problems are evaluated with both LLMs, in single-turn and multi-turn settings. The projection problem is evaluated only with \gem{} using iterative refinement with $n=5$. Therefore, projection problem results should be interpreted as a case study of the LLM's feasibility and optimality capabilities.

\subsubsection{Problem Instances}
To study scaling behaviour, we vary the number of channels over a wide range, depending on the LLM. In all problems, the total power is set proportional to the problem dimension $K$. For each $K$, we simulate different scenarios, sample random channel gains, and evaluate both models and baseline algorithms on the same set of instances. To better assess LLM optimization capabilities, we consider two cases: ($i$) \textit{single-turn} generation, i.e., no refinement, and ($ii$) \textit{multi-turn} iterative refinement. The detailed parameter grid can be found in Table~\ref{tab:exp_params}.
\begin{table*}[!t]
\centering
\caption{Experiment parameters for the representative optimization problems.}
\label{tab:exp_params}
\renewcommand{\arraystretch}{1.15}
\footnotesize
\begin{tabularx}{\textwidth}{@{} >{\raggedright\arraybackslash}p{0.21\textwidth} >{\raggedright\arraybackslash}p{0.20\textwidth} >{\raggedright\arraybackslash}X @{}}
\toprule
\textbf{Task} & \textbf{Parameter} & \textbf{Setting} \\
\midrule
\multirow{6}{*}{Sum-rate (Water-filling)}
& Channel count &
\begin{tabular}[t]{@{}l@{}}
OpenAI:\\
$K \in \{50,60,\ldots,140,150\}$ \\
Gemini:\\ $K \in \{80,90,\ldots,200,300,\ldots,700,800\}$
\end{tabular}
\\
& Gains & $h_i \sim \mathcal{CN}(0,1)$, $g_i = |h_i|^2$ \\
& Noise & $N_i = 1$ \\
& Power budget & $P_{\text{total}} = 0.4K$ \\
& Objective & $R(\mathbf{x}) = \sum_{i=1}^K \log_2\!\left(1 + \frac{x_i}{f_i}\right)$ \\
& Iteration per Scenario & $n \in \{1, 5\}$ \\
\midrule
\multirow{7}{*}{Energy efficiency (Fractional)}
& Channel count & 
\begin{tabular}[t]{@{}l@{}}
OpenAI:\\
$K \in \{40,50,\ldots,140,150\}$ \\
Gemini:\\ 
$K \in \{40,60,\ldots,180,200\}$
\end{tabular}\\
& Gains & $h_i \sim \mathcal{CN}(0,1)$, $g_i = |h_i|^2$ \\
& Noise & $N_i = 1$ \\
& Power budget & $P_{\text{total}} = 0.4K$ \\
& Circuit power & $P_{\text{const}} = 0.1K$ \\
& Objective & $\eta(\mathbf{x}) = \dfrac{R(\mathbf{x})}{P_{\text{const}} + \sum_{i=1}^K x_i}$ \\
& Iteration per Scenario & $n \in \{1, 5\}$ \\
\midrule
\multirow{6}{*}{Projection}
& Variables & $K \in \{30, 40, 50, 60\}$ \\
& Equality constraints & $p = 1$ \\
& Inequality constraints & $m = 1$ \\
& Equality tolerance & $\tau \in \{10^{-4}, 10^{-3}, 10^{-2}\}$ \\
& Objective & $\|\mathbf{x} - \mathbf{y}\|^2$ \\
& Iteration per Scenario & $n = 5$ \\
\bottomrule
\end{tabularx}
\end{table*}

\subsubsection{Experimental Protocol}
For each optimization family, model, and problem dimension, we evaluate the LLM on 30 randomly generated instances. For sum-rate and energy-efficiency maximization, we consider both a single-turn setting and a multi-turn setting, while for the projection problem we only consider the multi-turn setting. To better understand the improvement from iterative refinement, we test sum-rate maximization with a finer grid of iteration count $n$ using \gem{} and report the optimality rate for each $(K, n)$ pair.

Each LLM response is parsed and validated using rules defined in \reviewchange{Section~\ref{subsec:llm_protocol}} before the performance is scored. For valid solutions, we compute the objective value and compare it with the deterministic reference solution. Invalid responses are not treated as feasible candidates and are instead recorded as \textit{failure}. In the multi-turn setting, the best valid solution obtained within the allowed attempts is recorded for each instance. If no valid solution is obtained, the instance is counted as a \textit{failure}.

\subsection{Results on Optimization Capability}
We evaluate \omini{} and \gem{} on the two power-allocation problems. Both models produce valid solutions that are close to the deterministic optimum for small and moderate problem dimensions. Performance varies with the optimization family, model, and problem dimension $K$. For the projection problem, we only evaluate \gem{} in the iterative setting, since single-turn results does not yield enough valid outputs for a meaningful quality comparison.

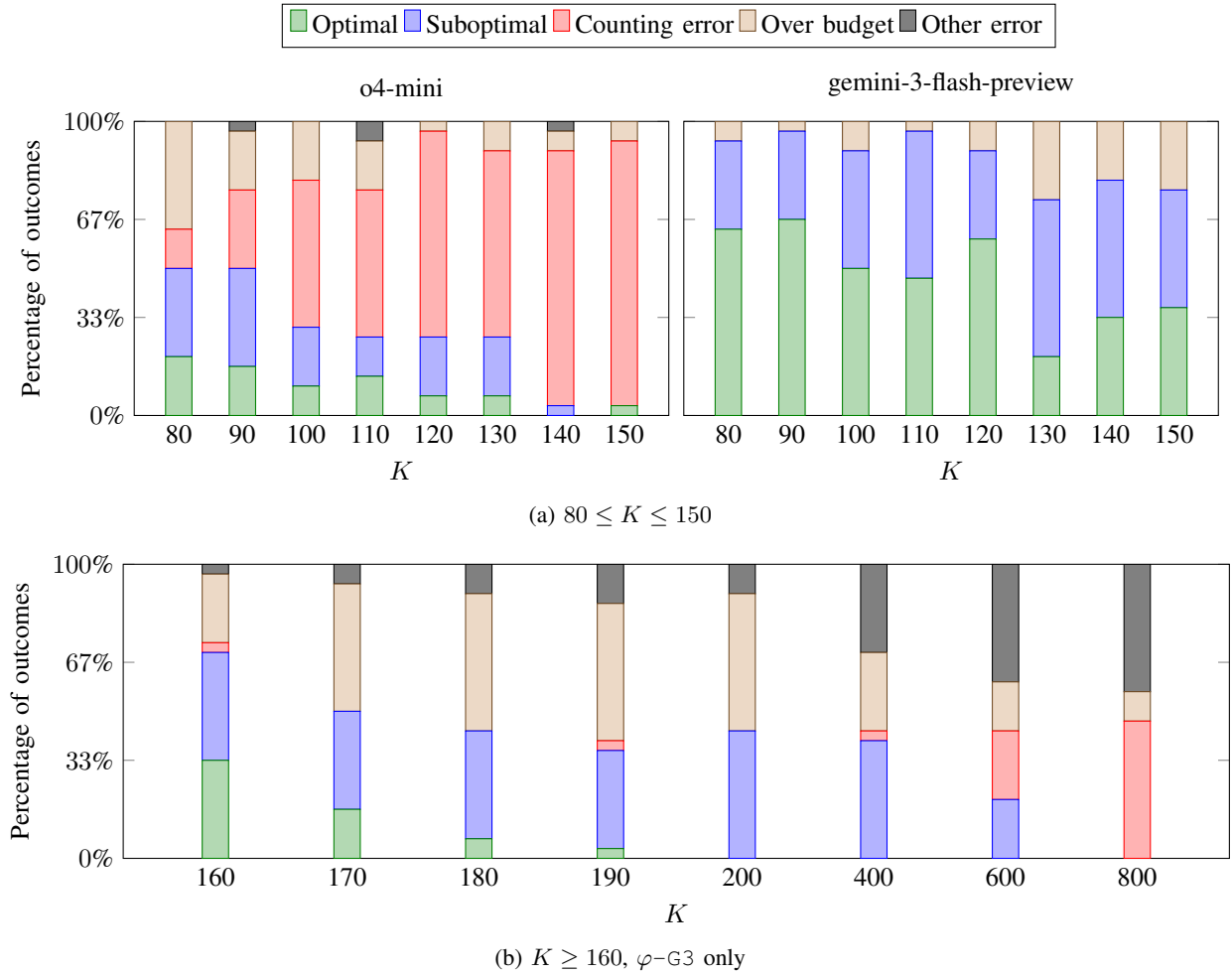
\begin{figure*}[!t]
\centering
\begin{subfigure}{\linewidth}
  \centering
  \begin{tikzpicture}
\begin{groupplot}[
  group style={group size=2 by 1, horizontal sep=0.2cm},
  width=0.48\linewidth,
  height=5.5cm,
  ybar stacked,
  ymin=0, ymax=30,
  xlabel={$K$},
  ylabel={Percentage of outcomes},
yticklabel={
    \pgfmathparse{\tick/30*100}
    \pgfmathprintnumber[fixed, precision=0]{\pgfmathresult}\%
},
  xtick=data,
  legend style={at={(1,1.25)}, anchor=south, legend columns=5},
]

% ---------- Top subplot ----------
\nextgroupplot[
    title={o4-mini},
  symbolic x coords={ 80, 90, 100, 110, 120, 130, 140, 150},
]
% optimal
\addplot[
  draw=successGreen, 
  fill=successGreen!30,
  forget plot,
  cycle list shift=-1
]  coordinates { (80,6) (90,5) (100,3) (110,4) (120,2) (130,2) (140,0) (150,1)};
\addlegendimage{ybar, draw=successGreen, fill=successGreen!30}
\addlegendentry{Optimal}
% suboptimal
\addplot coordinates {(80, 9) (90, 10) (100, 6) (110, 4) (120, 6) (130, 6) (140, 1) (150, 0)};
\addlegendentry{Suboptimal}
% counting error
\addplot coordinates {(80, 4) (90, 8) (100, 15) (110, 15) (120, 21) (130, 19) (140, 26) (150, 27)};
\addlegendentry{Counting error}
% over budget
\addplot coordinates {(80, 11) (90, 6) (100, 6) (110, 5) (120, 1) (130, 3) (140, 2) (150, 2)};
\addlegendentry{Over budget}
% other error
\addplot coordinates { (80, 0) (90, 1) (100, 0) (110, 2) (120, 0) (130, 0) (140, 1) (150, 0)};
\addlegendentry{Other error}

% ---------- Bottom subplot ----------
\nextgroupplot[
  title={gemini-3-flash-preview},
  xlabel={$K$}, 
  yticklabels=\empty,
  ylabel={},
  symbolic x coords={80, 90, 100, 110, 120, 130, 140, 150},
]
% optimal
\addplot[
  draw=successGreen, 
  fill=successGreen!30,
  forget plot,
  cycle list shift=-1
]  coordinates {(80,19) (90,20) (100,15) (110,14) (120,18) (130, 6) (140,10) (150,11) };
% suboptimal
\addplot coordinates {(80, 9) (90, 9) (100, 12) (110, 15) (120, 9) (130, 16) (140, 14) (150, 12) };
% counting error
\addplot coordinates {(80, 0) (90, 0) (100, 0) (110, 0) (120, 0) (130, 0) (140, 0) (150, 0)};
% over budget
\addplot coordinates {(80, 2) (90, 1) (100, 3) (110, 1) (120, 3) (130, 8) (140, 6) (150, 7)};
% other error
\addplot coordinates {(80, 0) (90, 0) (100, 0) (110, 0) (120, 0) (130, 0) (140, 0) (150, 0) };

\end{groupplot}
\end{tikzpicture}
  \caption{$80\le K \le 150$}
\end{subfigure}

\vspace{0.5\baselineskip}

\begin{subfigure}{\linewidth}
  \centering
  \begin{tikzpicture}
\begin{axis}[
  ybar stacked,
  width=0.9\linewidth,
  height=5.5cm,
  ymin=0, ymax=30,
  xlabel={$K$},
  ylabel={Percentage of outcomes},
yticklabel={
    \pgfmathparse{\tick/30*100}
    \pgfmathprintnumber[fixed, precision=0]{\pgfmathresult}\%
},
  xtick=data,
  symbolic x coords={160, 170, 180, 190, 200, 400, 600, 800},
]
% optimal
\addplot[
  draw=successGreen, 
  fill=successGreen!30,
  forget plot,
  cycle list shift=-1
] coordinates {(160, 10) (170,5) (180,2) (190, 1) (200,0) (400, 0) (600, 0) (800, 0)};
% \addlegendimage{ybar, draw=successGreen, fill=successGreen!30}
% \addlegendentry{Optimal}
% suboptimal
\addplot coordinates {(160, 11) (170, 10) (180, 11) (190, 10) (200, 13) (400, 12) (600, 6) (800, 0)};
% \addlegendentry{Suboptimal}
% counting error
\addplot coordinates { (160, 1) (170, 0) (180, 0) (190, 1) (200, 0) (400, 1) (600, 7) (800, 14)};
% \addlegendentry{Counting error}
% over budget
\addplot coordinates { (160,7) (170, 13) (180, 14) (190, 14) (200, 14) (400, 8) (600, 5) (800, 3)};
% \addlegendentry{Over budget}
% other error
\addplot coordinates { (160, 1) (170, 2) (180, 3) (190,4) (200, 3) (400, 9) (600, 12) (800, 13)};
% \addlegendentry{Other error}
\end{axis}
\end{tikzpicture}
  \caption{$K \ge 160$, \gem{} only}
\end{subfigure}
\caption{\revision{Distribution of optimization outcomes for sum-rate maximization without iterative refinement ($n=1$): (a) both models for $80\le K\le150$; (b) \gem{} only for $K\ge160$.}}
\label{fig:sum-rate-success-rate}
\end{figure*}

Results for single-turn sum-rate maximization are given in Figure~\ref{fig:sum-rate-success-rate}. As the figure shows, both models produce valid and near-optimal allocations at smaller problem dimensions. The proportions of valid and optimal solutions decrease as $K$ increases. In the single-turn setting, valid outputs at smaller $K$ frequently achieve the optimal or near-optimal objective value. \gem{} maintains consistent performance over a wider range of dimensions than \omini{}. At $K=90$, \omini{} produces valid solutions in approximately $50\%$ of the scenarios. The validity rate declines relatively slower for \gem{}. It remains above $50\%$ through $K=170$ and above $33.3\%$ through $K=400$. %Figure~\ref{fig:sum-rate-success-rate} summarizes these results, with 
Figure~\ref{fig:sum-rate-success-rate}(a) shows smaller $K$ ranges and Figure~\ref{fig:sum-rate-success-rate}(b) presents the results for larger $K$ ranges.

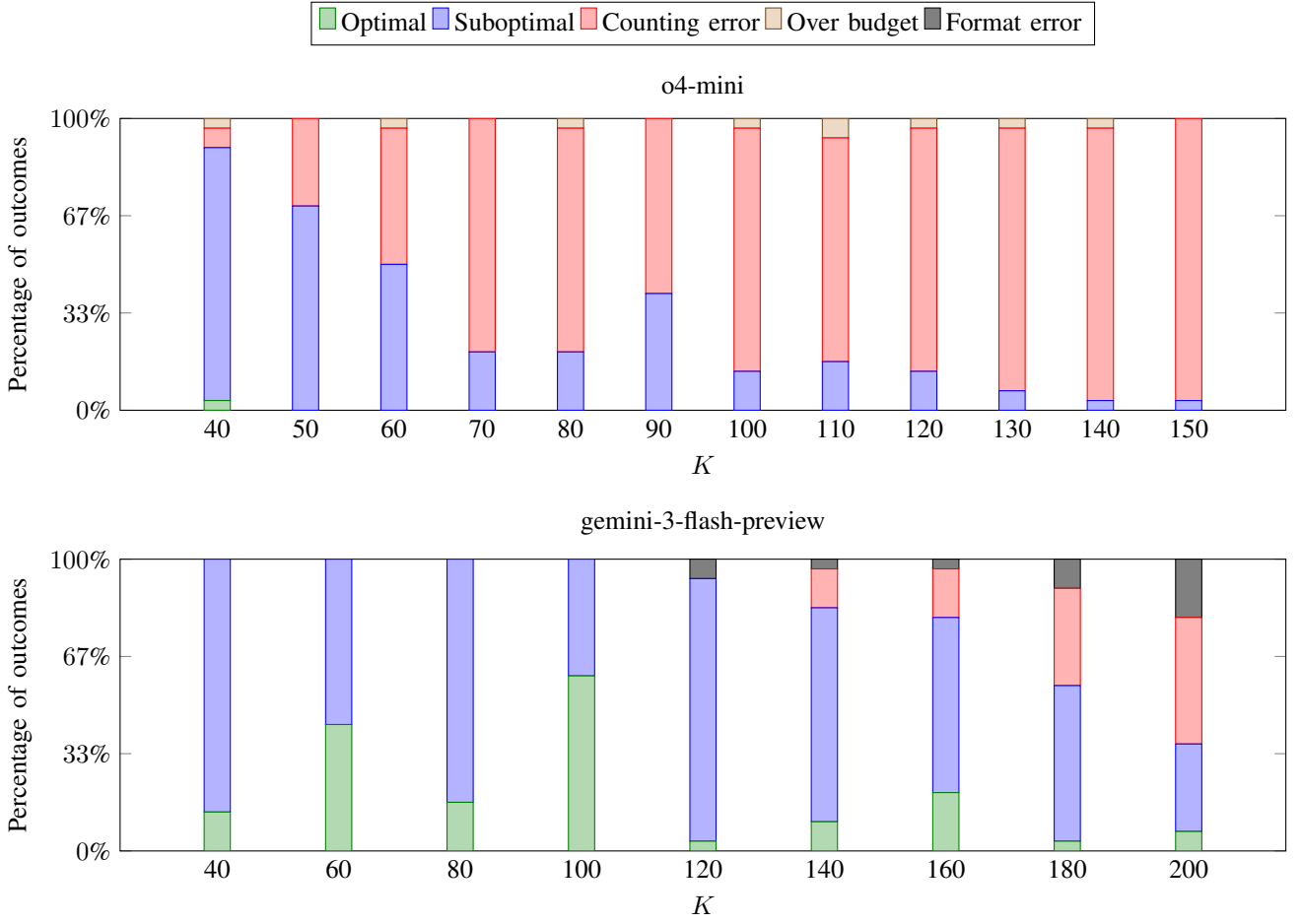
\begin{figure*}[!t]
\centering
\begin{tikzpicture}
\begin{groupplot}[
  group style={group size=1 by 2, vertical sep=2cm},
  width=0.95\linewidth,
  height=5.5cm,
  ybar stacked,
  ymin=0, ymax=30,
  xlabel={$K$},
  ylabel={Percentage of outcomes},
yticklabel={
    \pgfmathparse{\tick/30*100}
    \pgfmathprintnumber[fixed, precision=0]{\pgfmathresult}\%
},
  xtick=data,
  legend style={at={(0.5,1.25)}, anchor=south, legend columns=5},
]

% ---------- Top subplot ----------
\nextgroupplot[
    title={o4-mini},
  symbolic x coords={ 40,50,60,70,80, 90, 100, 110, 120, 130, 140, 150},
]
% optimal
\addplot[
  draw=successGreen, 
  fill=successGreen!30,
  forget plot,
  cycle list shift=-1
] coordinates { (40, 1) (50,0) (60,0) (70,0) (80, 0) (90, 0) (100, 0) (110, 0) (120, 0) (130, 0) (140, 0) (150, 0)};
\addlegendimage{ybar, draw=successGreen, fill=successGreen!30}
\addlegendentry{Optimal}
% suboptimal
\addplot coordinates {(40,26) (50,21) (60,15) (70,6) (80, 6) (90, 12) (100, 4) (110, 5) (120, 4) (130, 2) (140, 1) (150, 1)};
\addlegendentry{Suboptimal}
% counting error
\addplot coordinates {(40,2) (50,9) (60,14) (70,24) (80, 23) (90, 18) (100, 25) (110, 23) (120, 25) (130, 27) (140, 28) (150, 29)};
\addlegendentry{Counting error}
% over budget
\addplot coordinates {(40,1) (50,0) (60,1) (70,0) (80, 1) (90, 0) (100, 1) (110, 2) (120, 1) (130, 1) (140, 1) (150, 0)};
\addlegendentry{Over budget}
% other error
\addplot coordinates {(40,0) (50,0) (60,0) (70,0) (80, 0) (90, 0) (100, 0) (110, 0) (120, 0) (130, 0) (140, 0) (150, 0)};
\addlegendentry{Format error}
% \legend{Valid, Counting error, Over budget, Format error}

% ---------- Bottom subplot ----------
\nextgroupplot[
  title={gemini-3-flash-preview},
  xlabel={$K$}, 
  symbolic x coords={40, 60, 80, 100, 120, 140, 160, 180, 200},
]
% optimal
\addplot[
  draw=successGreen, 
  fill=successGreen!30,
  forget plot,
  cycle list shift=-1
] coordinates { (40, 4) (60,13)  (80, 5) (100, 18)  (120, 1)  (140, 3) (160, 6) (180, 1) (200, 2)};
% suboptimal
\addplot coordinates {(40, 26) (60, 17) (80, 25) (100, 12) (120, 27) (140, 22) (160,18) (180,16) (200,9) };
% counting error
\addplot coordinates {(40, 0) (60, 0) (80, 0) (100, 0) (120, 0) (140, 4) (160,5) (180,10) (200,13) };
% over budget
\addplot coordinates {(40, 0) (60, 0) (80, 0) (100, 0) (120, 0) (140, 0) (160,0) (180,0) (200,0) };
% other error
\addplot coordinates {(40, 0) (60, 0) (80, 0) (100, 0) (120, 2) (140, 1) (160,1) (180,3) (200,6) };

\end{groupplot}
\end{tikzpicture}
\caption{Distribution of LLM fractional programming results for single-turn energy efficiency maximization. Compared to convex optimization, fractional programming represents a more complicated structure, resulting in lower optimality rates.}
\label{fig:frac-outcome-distribution}
\end{figure*}

The energy-efficiency maximization task produces fewer optimal solutions than the sum-rate task. Although both models generate valid and occasionally optimal allocations at small $K$, the variance across sample instances is higher. For instance, \omini{} reaches optimality in approximately $3\%$ of scenarios at $K=40$. \gem{} produces optimal solutions at higher dimensions, but at lower rates than in sum-rate maximization. The results are given in Figure~\ref{fig:frac-outcome-distribution}.

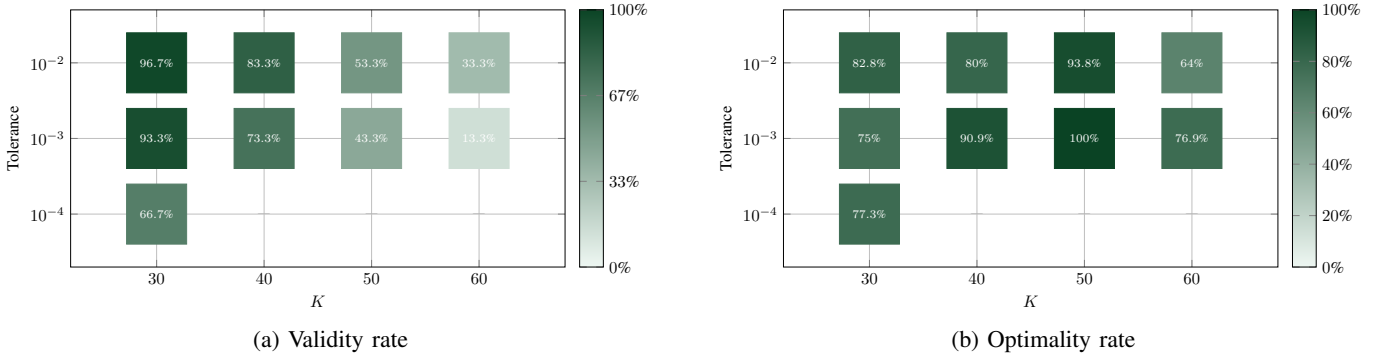
\begin{figure*}[!t]
\centering

\begin{subfigure}[t]{0.48\textwidth}
  \centering
  \resizebox{\linewidth}{!}{%
    % Heatmap: Projection validity rate vs K and tolerance
% Data from grid-20260419 results
% Native cells use original validation; derived cells re-evaluate
%   constraint violations against the looser tolerance.
% Usage: \input{figures/projection-validity-heatmap}
\begin{tikzpicture}
\begin{axis}[
  width=12cm, height=7cm,
  xlabel={$K$},
  ylabel={Tolerance},
  grid=both,
  colorbar,
  colormap name=greengradient,
  point meta min=0,
  point meta max=30,
  colorbar style={
    yticklabel={
        \pgfmathparse{\tick/30*100}
        \pgfmathprintnumber[fixed, precision=0]{\pgfmathresult}\%
    },
  },
  xtick={30,40,50,60},
  ytick={1,2,3},
  yticklabels={$10^{-4}$, $10^{-3}$, $10^{-2}$},
  xmin=22, xmax=68,
  ymin=0.3, ymax=3.7,
]
% --- Gray squares for cells without data ---
% \addplot[
%   only marks,
%   mark=square*,
%   mark size=18pt,
%   mark options={draw=gray!40, fill=gray!20},
% ] coordinates {
%   (40,1) (50,1) (60,1)
% };
% --- Colored squares for cells with data ---
\addplot[
  scatter,
  only marks,
  mark=square*,
  mark size=18pt,
  scatter src=explicit,
  scatter/use mapped color={draw=mapped color, fill=mapped color},
] table[x=x,y=y,meta=v] {
x   y    v
30  1    20
30  2    28
30  3    29
40  2    22
40  3    25
50  2    13
50  3    16
60  2    4
60  3    10
};
% --- Text labels (LaTeX computes v/30 * 100) ---
\addplot+ [
  only marks,
  mark=none,
  nodes near coords={
    \pgfmathparse{\pgfplotspointmeta/30*100}
    \pgfmathprintnumber[fixed, precision=1]{\pgfmathresult}\%
  },
  nodes near coords style={font=\scriptsize, anchor=center, text=white},
  point meta=explicit symbolic,
] table[x=x,y=y,meta=v] {
x   y    v
30  1    20
30  2    28
30  3    29
40  2    22
40  3    25
50  2    13
50  3    16
60  2    4
60  3    10
};
% --- Labels for cells without data ---
\node[font=\scriptsize, text=gray!60, anchor=center] at (axis cs:40,1) {---};
\node[font=\scriptsize, text=gray!60, anchor=center] at (axis cs:50,1) {---};
\node[font=\scriptsize, text=gray!60, anchor=center] at (axis cs:60,1) {---};
\end{axis}
\end{tikzpicture}%
  }
  \caption{Validity rate}
\end{subfigure}
\hfill
\begin{subfigure}[t]{0.48\textwidth}
  \centering
  \resizebox{\linewidth}{!}{%
    % Heatmap: Projection optimality rate (optimal/valid) vs K and tolerance
% Data from grid-20260419 results
% Optimality: obj_ratio >= 1 - 1e-6, rate = optimal_count / valid_count
% Usage: \input{figures/projection-optimality-heatmap}
\begin{tikzpicture}
\begin{axis}[
  width=12cm, height=7cm,
  xlabel={$K$},
  ylabel={Tolerance},
  grid=both,
  colorbar,
  colormap name=greengradient,
  point meta min=0,
  point meta max=100,
  colorbar style={
    yticklabel={
        \pgfmathprintnumber[fixed, precision=0]{\tick}\%
    },
  },
  xtick={30,40,50,60},
  ytick={1,2,3},
  yticklabels={$10^{-4}$, $10^{-3}$, $10^{-2}$},
  xmin=22, xmax=68,
  ymin=0.3, ymax=3.7,
]
% --- Gray squares for cells without data ---
% \addplot[
%   only marks,
%   mark=square*,
%   mark size=18pt,
%   mark options={draw=gray!40, fill=gray!20},
% ] coordinates {
%   (40,1) (50,1) (60,1)
% };
% --- Colored squares for cells with data ---
% v = optimal/valid * 100 (pre-computed since denominators vary)
% K=30: 17/22=77.3, 21/28=75.0, 24/29=82.8
% K=40: 20/22=90.9, 20/25=80.0
% K=50: 13/13=100, 15/16=93.8
% K=60: 10/13=76.9, 16/25=64.0
\addplot[
  scatter,
  only marks,
  mark=square*,
  mark size=18pt,
  scatter src=explicit,
  scatter/use mapped color={draw=mapped color, fill=mapped color},
] table[x=x,y=y,meta=v] {
x   y    v
30  1    77.3
30  2    75.0
30  3    82.8
40  2    90.9
40  3    80.0
50  2    100.0
50  3    93.8
60  2    76.9
60  3    64.0
};
% --- Text labels ---
\addplot+ [
  only marks,
  mark=none,
  nodes near coords={
    \pgfmathprintnumber[fixed, precision=1]{\pgfplotspointmeta}\%
  },
  nodes near coords style={font=\scriptsize, anchor=center, text=white},
  point meta=explicit symbolic,
] table[x=x,y=y,meta=v] {
x   y    v
30  1    77.3
30  2    75.0
30  3    82.8
40  2    90.9
40  3    80.0
50  2    100.0
50  3    93.8
60  2    76.9
60  3    64.0
};
% --- Labels for cells without data ---
\node[font=\scriptsize, text=gray!60, anchor=center] at (axis cs:40,1) {---};
\node[font=\scriptsize, text=gray!60, anchor=center] at (axis cs:50,1) {---};
\node[font=\scriptsize, text=gray!60, anchor=center] at (axis cs:60,1) {---};
\end{axis}
\end{tikzpicture}%
  }
  \caption{Optimality rate}
\end{subfigure}

\caption{Heatmap of projection problem validity and optimality rates
across different variable counts ($K$) and equality-constraint
tolerance levels, using \gem{}.}
\label{fig:projection-heatmap}
\end{figure*}

The projection task provides an additional test beyond the two power-allocation formulations. As shown in Figure~\ref{fig:projection-heatmap}, validity decreases with increasing $K$, while relaxing the equality-constraint tolerance improves validity. Among valid outputs, the model achieves high optimality rates across the evaluated $(K,\tau)$ configurations. Compared with sum-rate maximization, the LLM-based solver shows a degraded performance in this case. Noting that both problems belong to convex optimization family, this observation suggests that LLM is aided by the known structure of sum-rate maximization task. 

\subsection{Results on Scaling Behaviour}
Across all problems, validity and optimality rates decrease as the problem dimension increases. However, the two measures scale differently. For sum-rate maximization, the single-turn validity rate of \omini{} decreases to approximately $50\%$ at $K=90$. In comparison, \gem{} remains valid more than $50\%$ up to approximately $K=170$ and continues to return valid outputs beyond $K=200$. The end-to-end optimality rate of \gem{} decreases substantially at larger $K$, with no optimal solutions observed when $K\geq190$, as shown in Figure~\ref{fig:sum-rate-success-rate}(b).

The energy-efficiency task shows a \textit{narrower} reliable operating range than sum-rate maximization, which is intuitive due to the higher complexity of underlying optimization. Both models experience counting and validity failures at lower dimensions, and the optimality rate drops noticeably as compared with sum-rate maximization.

The projection results show a similar scaling trend. With $\tau = 10^{-3}$, the validity rate decreases from $93.3\%$ at $K=30$ to $13.3\%$ at $K=60$. Relaxing the equality-constraint tolerance improves validity by a shift on the $K$-axis, but it does not meaningfully impact the slope of drop in the probability of generating a feasible solution.

\subsection{Failure Modes}

The observed failures vary across models, optimization families, and problem dimensions. At smaller dimensions, overbudget allocations are a common failure mode in the two power-allocation tasks. These outputs often resemble the allocation pattern of the deterministic solution but exceed the total-power constraint by a small amount; see \comparisonlocation{}.

As the problem dimension increases, counting errors become the dominant failure mode. In these cases, the model returns a numerical list whose length does not match the required dimension. This failure becomes particularly prevalent for \omini{} in sum-rate maximization when $K\geq100$. \gem{} maintains correct output length over a wider range, but also exhibits counting failures at sufficiently large dimensions.

Format violations are more prominent for \gem{} at very large $K$. Rather than returning a numerical list in the requested JSON format, the model may include additional text or alternative delimiters, which is a behaviour we have not observed for \omini{}. For the projection task, numerical constraint violations are the main source of invalidity. Candidate solutions may be close to feasible but fail the equality constraint $A\mathbf{x}+\mathbf{b}=\mathbf{0}$ under a strict tolerance.

\subsection{Effect of Iterative Refinement}
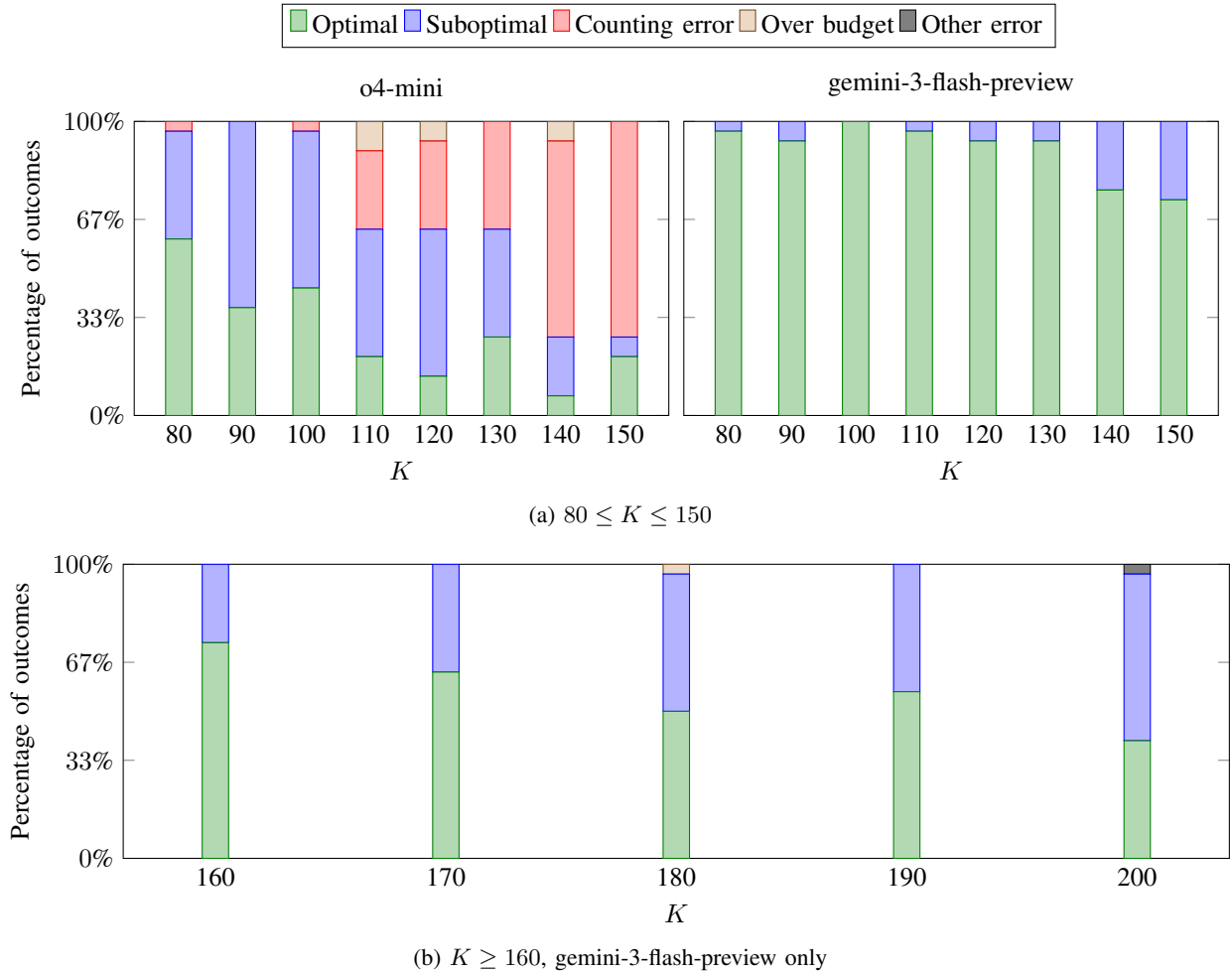
\begin{figure*}[h]
\centering
\begin{subfigure}{\linewidth}
  \centering
  \begin{tikzpicture}
\begin{groupplot}[
  group style={group size=2 by 1, horizontal sep=0.2cm},
  width=0.48\linewidth,
  height=5.5cm,
  ybar stacked,
  ymin=0, ymax=30,
  xlabel={$K$},
  ylabel={Percentage of outcomes},
yticklabel={
    \pgfmathparse{\tick/30*100}
    \pgfmathprintnumber[fixed, precision=0]{\pgfmathresult}\%
},
  xtick=data,
  legend style={at={(1,1.25)}, anchor=south, legend columns=5},
]

% ---------- Top subplot ----------
\nextgroupplot[
    title={o4-mini},
  symbolic x coords={80, 90, 100, 110, 120, 130, 140, 150},
]
% optimal
\addplot[
  draw=successGreen, 
  fill=successGreen!30,
  forget plot,
  cycle list shift=-1
] coordinates {(80,18) (90,11) (100,13) (110,6) (120,4) (130,8) (140,2) (150,6)};
\addlegendimage{ybar, draw=successGreen, fill=successGreen!30}
\addlegendentry{Optimal}
% suboptimal
\addplot coordinates {(80, 11) (90, 19) (100, 16) (110, 13) (120, 15) (130, 11) (140, 6) (150, 2)};
\addlegendentry{Suboptimal}
% counting error
\addplot coordinates {(80, 1) (90, 0) (100, 1) (110, 8) (120, 9) (130, 11) (140, 20) (150, 22)};
\addlegendentry{Counting error}
% over budget
\addplot coordinates {(80, 0) (90, 0) (100, 0) (110, 3) (120, 2) (130, 0) (140, 2) (150, 0)};
\addlegendentry{Over budget}
% other error
\addplot coordinates {(80, 0) (90, 0) (100, 0) (110, 0) (120, 0) (130, 0) (140, 0) (150, 0)};
\addlegendentry{Other error}

% ---------- Bottom subplot ----------
\nextgroupplot[
  title={gemini-3-flash-preview},
  xlabel={$K$}, 
  yticklabels=\empty,
  ylabel={},
  symbolic x coords={ 80, 90, 100, 110, 120, 130, 140, 150},
]
% optimal
\addplot[
  draw=successGreen, 
  fill=successGreen!30,
  forget plot,
  cycle list shift=-1
] coordinates {(80, 29) (90, 28) (100, 30) (110, 29) (120, 28) (130, 28) (140, 23) (150, 22) };
% suboptimal
\addplot coordinates {(80, 1) (90, 2) (100, 0) (110, 1) (120, 2) (130, 2) (140, 7) (150, 8) };
% counting error
\addplot coordinates {(80, 0) (90, 0) (100, 0) (110, 0) (120, 0) (130, 0) (140, 0) (150, 0)};
% over budget
\addplot coordinates {(80, 0) (90, 0) (100, 0) (110, 0) (120, 0) (130, 0) (140, 0) (150, 0)};
% other error
\addplot coordinates {(80, 0) (90, 0) (100, 0) (110, 0) (120, 0) (130, 0) (140, 0) (150, 0)};

\end{groupplot}
\end{tikzpicture}
  \caption{$80\le K \le 150$}
\end{subfigure}

\vspace{0.5\baselineskip}

\begin{subfigure}{\linewidth}
  \centering
  \begin{tikzpicture}
\begin{axis}[
  ybar stacked,
  width=0.9\linewidth,
  height=5.5cm,
  ymin=0, ymax=30,
  xlabel={$K$},
  ylabel={Percentage of outcomes},
yticklabel={
    \pgfmathparse{\tick/30*100}
    \pgfmathprintnumber[fixed, precision=0]{\pgfmathresult}\%
},
  xtick=data,
  symbolic x coords={160, 170, 180, 190, 200},
]
% optimal
\addplot[
  draw=successGreen, 
  fill=successGreen!30,
  forget plot,
  cycle list shift=-1
] coordinates {(160, 22) (170,19) (180,15) (190, 17) (200,12)};
% \addlegendimage{ybar, draw=successGreen, fill=successGreen!30}
% \addlegendentry{Optimal}
% suboptimal
\addplot coordinates {(160, 8) (170, 11) (180, 14) (190, 13) (200, 17) };
% \addlegendentry{Suboptimal}
% counting error
\addplot coordinates { (160, 0) (170, 0) (180, 0) (190, 0) (200, 0) };
% \addlegendentry{Counting error}
% over budget
\addplot coordinates { (160,0) (170, 0) (180, 1) (190, 0) (200, 0) };
% \addlegendentry{Over budget}
% other error
\addplot coordinates { (160, 0) (170, 0) (180, 0) (190,0) (200, 1)};
% \addlegendentry{Other error}
\end{axis}
\end{tikzpicture}
  \caption{$K \ge 160$, gemini-3-flash-preview only}
\end{subfigure}
\caption{Distribution of LLM optimization results for sum-rate maximization with iterative refinement ($n=5$). Compared to the single-turn setting, providing targeted numerical feedback substantially recovers format validity and pushes the optimality boundary to larger $K$ domains.}
\label{fig:sum-rate-iterative}
\end{figure*}

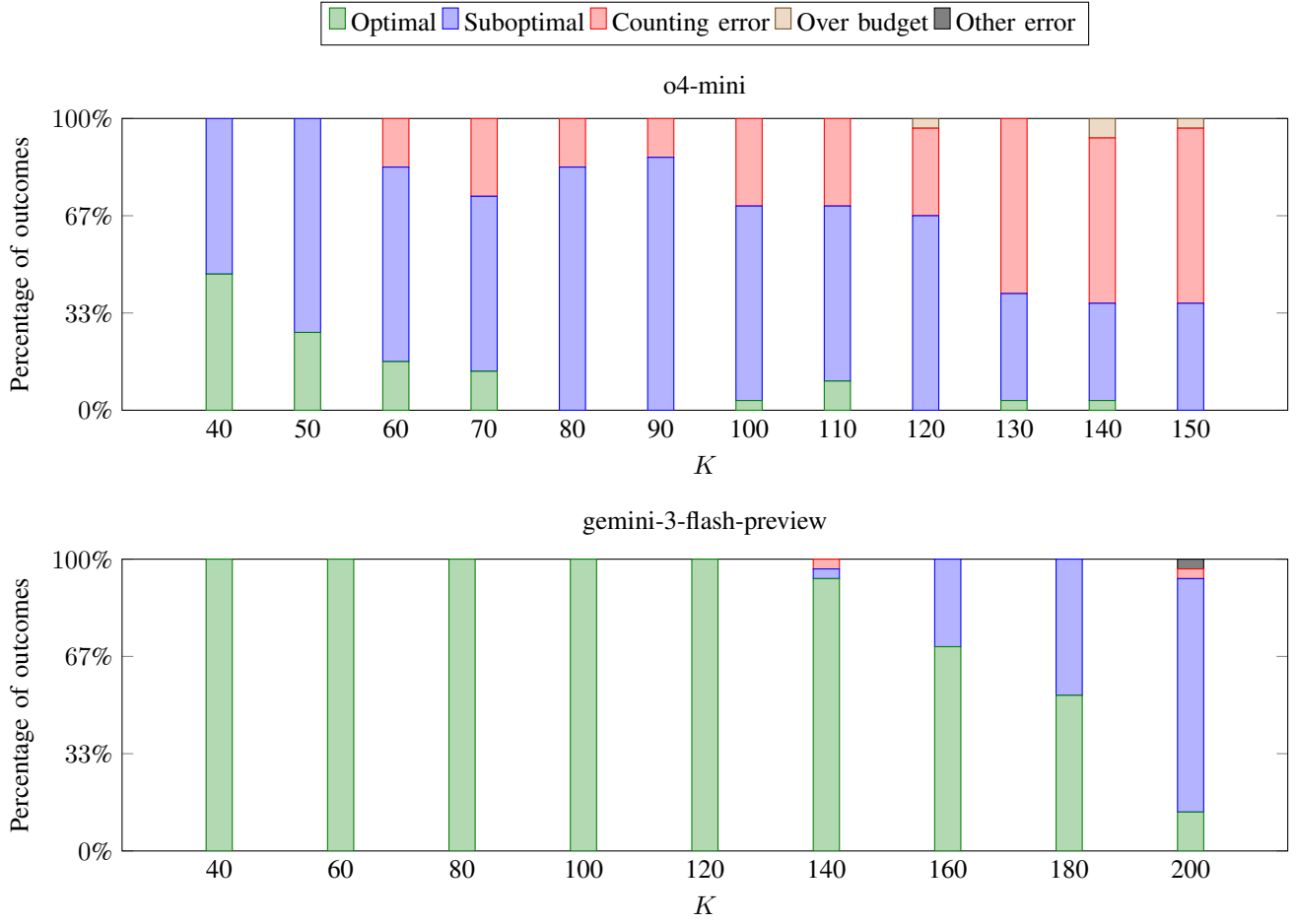
\begin{figure*}[h]
\centering
\begin{tikzpicture}
\begin{groupplot}[
  group style={group size=1 by 2, vertical sep=2cm},
  width=0.95\linewidth,
  height=5.5cm,
  ybar stacked,
  ymin=0, ymax=30,
  xlabel={$K$},
  ylabel={Percentage of outcomes},
yticklabel={
    \pgfmathparse{\tick/30*100}
    \pgfmathprintnumber[fixed, precision=0]{\pgfmathresult}\%
},
  xtick=data,
  legend style={at={(0.5,1.25)}, anchor=south, legend columns=5},
]

% ---------- Top subplot ----------
\nextgroupplot[
    title={o4-mini},
  symbolic x coords={40,50,60,70,80, 90, 100, 110, 120, 130, 140, 150},
]
% optimal
\addplot[
  draw=successGreen, 
  fill=successGreen!30,
  forget plot,
  cycle list shift=-1
] coordinates { (40, 14) (50,8) (60,5) (70,4) (80,0) (90,0) (100,1) (110,3) (120,0) (130,1) (140,1) (150,0)};
\addlegendimage{ybar, draw=successGreen, fill=successGreen!30}
\addlegendentry{Optimal}
% suboptimal
\addplot coordinates {(40,16) (50,22) (60,20) (70, 18) (80, 25) (90, 26) (100, 20) (110, 18) (120, 20) (130, 11) (140, 10) (150, 11)};
\addlegendentry{Suboptimal}
% counting error
\addplot coordinates { (40, 0) (50,0) (60,5) (70,8) (80, 5) (90, 4) (100, 9) (110, 9) (120, 9) (130, 18) (140, 17) (150, 18)};
\addlegendentry{Counting error}
% over budget
\addplot coordinates { (40, 0) (50,0) (60,0) (70,0) (80, 0) (90, 0) (100, 0) (110, 0) (120, 1) (130, 0) (140, 2) (150, 1)};
\addlegendentry{Over budget}
% other error
\addplot coordinates {
  (40, 0) (50,0) (60,0) (70,0) (80, 0) (90, 0) (100, 0) (110, 0) (120, 0) (130, 0) (140, 0) (150, 0)};
\addlegendentry{Other error}
% \legend{Suboptimal, Counting error, Over budget, Other error, Optimal}

% ---------- Bottom subplot ----------
\nextgroupplot[
  title={gemini-3-flash-preview},
  xlabel={$K$}, 
  symbolic x coords={40,60,80, 100, 120, 140, 160, 180, 200},
]
% optimal
\addplot[
  draw=successGreen, 
  fill=successGreen!30,
  forget plot,
  cycle list shift=-1
] coordinates {(40, 30) (60, 30) (80, 30)  (100, 30) (120, 30) (140, 28) (160,21) (180,16) (200,4)};
% suboptimal
\addplot coordinates {(40, 0) (60, 0) (80, 0) (100, 0) (120, 0) (140, 1) (160,9) (180,14) (200,24)};
% counting error
\addplot coordinates {(40, 0) (60, 0) (80, 0) (100, 0) (120, 0) (140, 1) (160,0) (180,0) (200,1)};
% over budget
\addplot coordinates {(40, 0) (60, 0) (80, 0) (100, 0) (120, 0) (140, 0) (160,0) (180,0) (200,0)};
% other error
\addplot coordinates {(40, 0) (60, 0) (80, 0) (100, 0) (120, 0) (140, 0) (160,0) (180,0) (200,1)};

\end{groupplot}
\end{tikzpicture}
\caption{Distribution of LLM fractional programming results for power efficiency maximization leveraging iterative refinement ($n=5$). Providing boundary feedback actively mitigates structural errors, helping both models explore fractional spaces more efficiently.}
\label{fig:frac-iterative-outcome}
\end{figure*}

We compare the single-turn setting with a multi-turn refinement setting that allows up to five attempts. Figures~\ref{fig:sum-rate-iterative} and \ref{fig:frac-iterative-outcome} show the resulting outcome distributions for sum-rate and energy-efficiency maximization, respectively.

For sum-rate maximization, allowing up to five attempts extends the range of dimensions for which valid solutions can be obtained. With refinement, \omini{} maintains a validity rate above $50\%$ through $K=130$, whereas \gem{} maintains validity above $90\%$ at $K=200$.

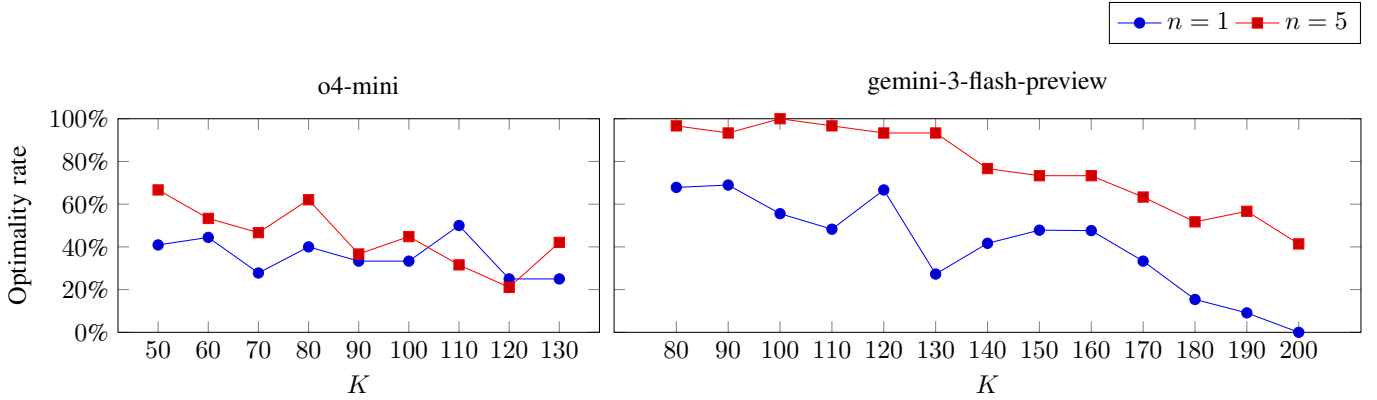
\begin{figure*}[h]
\centering
\vspace{1.5em}
\resizebox{\textwidth}{!}{\begin{tikzpicture}
\begin{groupplot}[
  group style={group size=2 by 1, horizontal sep=0.2cm},
  height=4.5cm,
  xlabel={$K$},
  ylabel={Optimality rate},
  xtick=data,
  ymax=1,
  ymin=0,
  yticklabel={
      \pgfmathparse{\tick*100}
      \pgfmathprintnumber[fixed, precision=0]{\pgfmathresult}\%
  },
  legend style={
    at={(axis description cs:1,1.35)},
    anchor=south east,
    legend columns=2,
    /tikz/overlay,
  },
]

% ---------- Top subplot ----------
\nextgroupplot[
    title={o4-mini},
    width=0.45\linewidth,
]
  \addplot coordinates {
    (50,9/22)
    (60,12/27)
    (70,5/18)
    (80,6/15)
    (90,5/15)
    (100,3/9)
    (110,4/8)
    (120,2/8)
    (130,2/8)
  };

  \addplot coordinates {
    (50,20/30)
    (60, 16/30)
    (70, 14/30)
    (80, 18/29)
    (90, 11/30)
    (100, 13/29)
    (110, 6/19)
    (120, 4/19)
    (130, 8/19)
  };

% ---------- Bottom subplot ----------
\nextgroupplot[
  title={gemini-3-flash-preview},
  xlabel={$K$}, 
  yticklabels=\empty,
  ylabel={},
    width=0.65\linewidth,
]
  \addplot coordinates {
    (80, 19/28)
(90, 20/29)
(100, 15/27)
(110, 14/29)
(120, 18/27)
(130, 6/22)
(140, 10/24)
(150, 11/23)
(160, 10/21)
(170, 5/15)
(180,2/13)
(190, 1/11)
(200,0/13)
  };
  \addlegendentry{$n=1$}

  \addplot coordinates {
    (80, 29/30)
    (90, 28/30)
    (100,30/30)
    (110, 29/30)
    (120, 28/30)
    (130, 28/30)
    (140, 23/30)
    (150, 22/30)
    (160, 22/30)
    (170, 19/30)
    (180, 15/29)
    (190, 17/30)
    (200, 12/29)
  };
  \addlegendentry{$n=5$}

\end{groupplot}
\end{tikzpicture}}
\caption{Optimality rate in the sum-rate maximization task. Applying 5 refinement iterations improves this rate across all $K$ values.}
\label{fig:sum-rate-optimal-ratio}
\end{figure*}

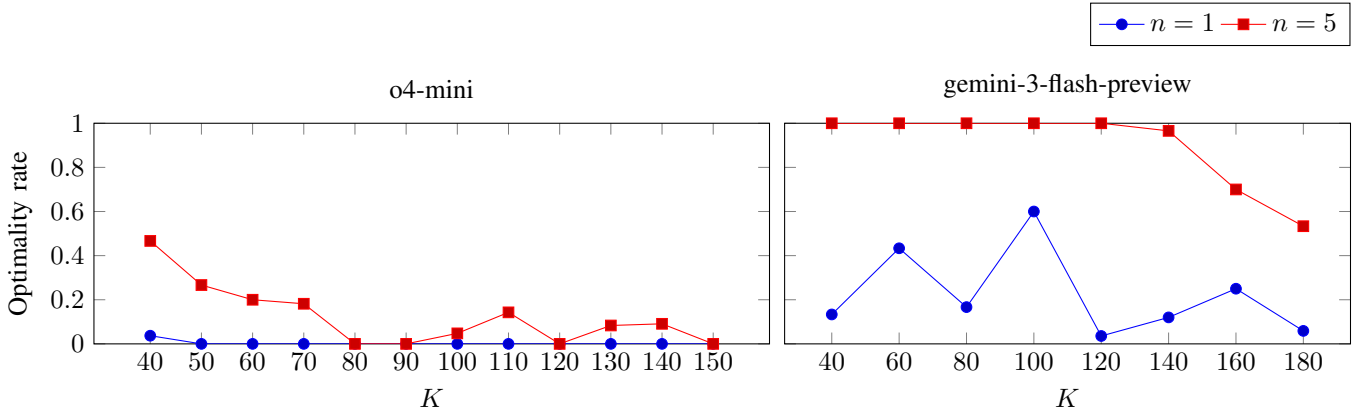
\begin{figure*}[h]
\centering
\vspace{1.5em}
\begin{tikzpicture}
\begin{groupplot}[
  group style={group size=2 by 1, horizontal sep=0.2cm},
  height=4.5cm,
  xlabel={$K$},
  ylabel={Optimality rate},
  xtick=data,
  ymax=1,
  ymin=0,
  legend style={
    at={(axis description cs:1,1.35)},
    anchor=south east,
    legend columns=2,
    /tikz/overlay,
  },
]

% ---------- Top subplot ----------
\nextgroupplot[
    title={o4-mini},
    width=0.58\linewidth,
]
  \addplot coordinates {
    (40, 1/27)
    (50, 0)
    (60, 0)
    (70, 0)
    (80, 0)
    (90, 0)
    (100, 0)
    (110, 0)
    (120, 0)
    (130, 0)
    (140, 0)
    (150, 0)
  };

  \addplot coordinates {
    (40, 14/30)
    (50, 8/30)
    (60, 5/25)
    (70, 4/22)
    (80, 0)
    (90, 0)
    (100, 1/21)
    (110, 3/21)
    (120, 0)
    (130, 1/12)
    (140, 1/11)
    (150, 0)
  };

% ---------- Bottom subplot ----------
\nextgroupplot[
  title={gemini-3-flash-preview},
  xlabel={$K$}, 
  yticklabels=\empty,
  ylabel={},
    width=0.5\linewidth,
]
  \addplot coordinates {
    (40, 4/30)
    (60, 13/30)
    (80, 5/30)
    (100, 18/30)
    (120, 1/28)
    (140, 3/25)
    (160, 6/24)
    (180, 1/17)
  };
  \addlegendentry{$n=1$}

  \addplot coordinates {
    (40, 1)
    (60, 1)
    (80, 1)
    (100,1)
    (120, 1)
    (140, 28/29)
    (160, 21/30)
    (180, 16/30)
  };
  \addlegendentry{$n=5$}

\end{groupplot}
\end{tikzpicture}
\caption{Optimality rate in the power efficiency task. Iterative refinement pushes the models to discover higher-quality solutions, with a more pronounced improvement for \gem{}.}
\label{fig:frac-optimal-ratio}
\end{figure*}

\begin{figure*}[h]
\centering
\begin{tikzpicture}
\begin{axis}[
  width=12cm, height=5.5cm,
  xlabel={$K$}, ylabel={iteration},
  grid=both,
  colorbar,
  colormap name=greengradient,
  point meta min=0,
  point meta max=30,
  colorbar style={
    yticklabel={
        \pgfmathparse{\tick/30*100}
        \pgfmathprintnumber[fixed, precision=0]{\pgfmathresult}\%
    },
  },
  legend style={at={(0.7,1.15)}, anchor=south},
]
\addplot[
  scatter,
  only marks,
  mark=square*,
  mark size=8pt,
  scatter src=explicit,
  scatter/use mapped color={draw=mapped color, fill=mapped color},
] table[x=x,y=y,meta=v] {
x   y    v
100   1  15
110   1   14
120  1   18
130 1 6
170 5 19
180  5  15
190  5  17
200  5  12
150  4  18
160  4  18
170  4  13
140  3  17
150 3 19
160 3 17
170 3 18
120 2 15
130 2 16
140 2 18
};
\addplot+ [
  only marks,
  mark=none,
  nodes near coords={
    \pgfmathparse{\pgfplotspointmeta/30*100}
    \pgfmathprintnumber[fixed, precision=0]{\pgfmathresult}\%
  },
  nodes near coords style={font=\tiny, anchor=center, text=white},
  point meta=explicit symbolic,
] table[x=x,y=y,meta=v] {
x   y    v
100   1  15
110   1   14
120  1   18
130 1 6
170 5 19
180  5  15
190  5  17
200  5  12
150  4  18
160  4  18
170  4  13
140  3  17
150 3 19
160 3 17
170 3 18
120 2 15
130 2 16
140 2 18
};

\end{axis}
\end{tikzpicture}
\caption{Heatmap of the end-to-end optimality rate generated by \gem{} for sum-rate maximization across different problem dimensions ($K$) and refinement iterations ($n$). Deeper colour intensities denote higher optimality ratios. }% of  configurations among valid outputs.}
\label{fig:sum-rate-success-rate-50-scatter}
\end{figure*}
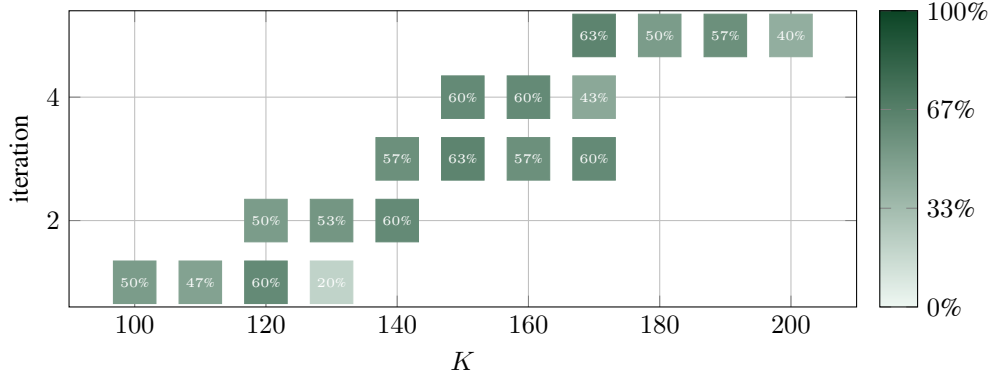

Refinement also improves solution quality among valid outputs. As shown in Figures~\ref{fig:sum-rate-optimal-ratio} and \ref{fig:frac-optimal-ratio}, increasing the number of attempts raises the proportion of valid solutions that achieve the deterministic optimum. The refinement heatmap in Figure~\ref{fig:sum-rate-success-rate-50-scatter} further reports the optimal-solution ratio of \gem{} across problem dimensions and iteration counts.

Across both power-allocation tasks, the multi-turn setting produces higher validity and optimality rates than the single-turn baseline. The improvement is larger for \gem{}, particularly at higher problem dimensions.

\subsection{Generalization Across Optimization Families}

Performance differs across the three optimization families. Sum-rate maximization supports the widest range of valid and high-quality LLM outputs. Energy-efficiency maximization exhibits lower end-to-end optimality rates and earlier validity failures as $K$ increases. For the projection task, valid outputs achieve high optimality rates, while feasibility remains sensitive to the equality-constraint tolerance.

The comparison across problems suggests that LLM performance depends on both problem dimension and mathematical structure. Sum-rate maximization is the most accessible of the evaluated tasks, where its concave objective and simple total-power constraint allow both models to produce valid, high-quality allocations over the widest range of dimensions.

Energy-efficiency maximization is more difficult due to its fractional objective. The ratio structure requires the model to balance rate improvement against total power consumption, and the resulting solutions are less consistently optimal even when they are valid. The earlier onset of counting errors and the lower optimality rates indicate that LLM performance is sensitive to the complexity of the underlying optimization structure, rather than being determined by dimension alone.

The projection task provides complementary evidence of generalization. Although it does not use a familiar power-allocation objective, \gem{} produces near-optimal projections whenever it satisfies the constraints. This suggests that LLMs \textit{can reason about optimization structures} beyond the specific named formulations, while the bottleneck remains the ability to generate valid outputs that meet problem constraints strictly.

\subsection{Computational Cost}

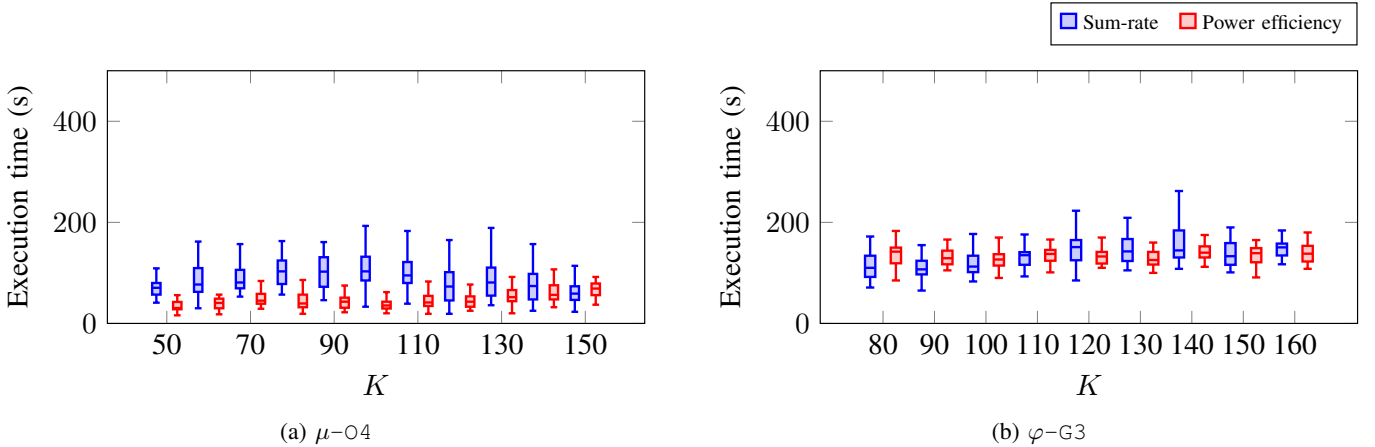
\begin{figure*}[!t]
\centering
\vspace{1.5em}

\begin{subfigure}[t]{0.48\textwidth}
  \centering
  \resizebox{\linewidth}{!}{%
    \begin{tikzpicture}
\begin{axis}[
    width=8cm,
    height=4.6cm,
    ymin=0,
    ymax=500,
    xlabel={$K$},
    ylabel={Execution time (s)},
    boxplot/draw direction=y,
    boxplot={
        box extend=0.2
    },
    xtick={1,3,5,7,9,11},
    xticklabels={50,70,90,110,130,150},
    every axis plot/.append style={thick},
    sumrate/.style={draw=blue, fill=blue!20, solid},
    power/.style={draw=red, fill=red!20, solid},
]

% k = 50
\addplot[
    boxplot,
    sumrate,
    boxplot/draw position=0.75,
] table[row sep=\\, y index=0] {
data \\
74	\\
72	\\
107	\\
82	\\
48	\\
181	\\
62	\\
59	\\
82	\\
61	\\
89	\\
63	\\
46	\\
64	\\
51	\\
65	\\
54	\\
41	\\
78	\\
41	\\
73	\\
129	\\
76	\\
55	\\
109	\\
69	\\
76	\\
73	\\
94	\\
};
\addplot[
    boxplot,
    power,
    boxplot/draw position=1.25,
] table[row sep=\\, y index=0] {
data \\
56	\\
36	\\
16	\\
47	\\
42	\\
24	\\
29	\\
46	\\
44	\\
28	\\
18	\\
28	\\
31	\\
54	\\
33	\\
30	\\
26	\\
100	\\
29	\\
42	\\
32	\\
28	\\
43	\\
49	\\
28	\\
25	\\
28	\\
42	\\
30	\\
};

% k = 60
\addplot[
    boxplot,
    sumrate,
    boxplot/draw position=1.75,
] table[row sep=\\, y index=0] {
data \\
51	\\
100	\\
30	\\
80	\\
62	\\
76	\\
74	\\
115	\\
162	\\
132	\\
110	\\
62	\\
99	\\
87	\\
105	\\
73	\\
117	\\
52	\\
75	\\
109	\\
133	\\
49	\\
78	\\
149	\\
64	\\
63	\\
116	\\
65	\\
52	\\
};
\addplot[
    boxplot,
    power,
    boxplot/draw position=2.25,
] table[row sep=\\, y index=0] {
data \\
50	\\
40	\\
28	\\
31	\\
18	\\
138	\\
24	\\
38	\\
207	\\
41	\\
96	\\
32	\\
50	\\
28	\\
28	\\
41	\\
57	\\
29	\\
38	\\
46	\\
47	\\
43	\\
117	\\
24	\\
44	\\
96	\\
40	\\
48	\\
33	\\
};

% k = 70
\addplot[
    boxplot,
    sumrate,
    boxplot/draw position=2.75,
] table[row sep=\\, y index=0] {
data \\
68	\\
91	\\
73	\\
71	\\
77	\\
82	\\
73	\\
66	\\
68	\\
107	\\
131	\\
122	\\
82	\\
57	\\
88	\\
57	\\
100	\\
269	\\
53	\\
102	\\
105	\\
113	\\
157	\\
126	\\
80	\\
76	\\
53	\\
150	\\
79	\\
};
\addplot[
    boxplot,
    power,
    boxplot/draw position=3.25,
] table[row sep=\\, y index=0] {
data \\
119	\\
39	\\
38	\\
38	\\
31	\\
53	\\
37	\\
41	\\
29	\\
38	\\
44	\\
101	\\
39	\\
65	\\
84	\\
39	\\
58	\\
41	\\
69	\\
39	\\
53	\\
48	\\
59	\\
42	\\
79	\\
46	\\
55	\\
59	\\
55	\\
};

% k = 80
\addplot[
    boxplot,
    sumrate,
    boxplot/draw position=3.75,
] table[row sep=\\, y index=0] {
data \\
90	\\
163	\\
104	\\
82	\\
69	\\
94	\\
57	\\
110	\\
102	\\
113	\\
125	\\
101	\\
92	\\
223	\\
81	\\
121	\\
66	\\
115	\\
74	\\
129	\\
75	\\
131	\\
124	\\
107	\\
70	\\
224	\\
75	\\
146	\\
130	\\
};
\addplot[
    boxplot,
    power,
    boxplot/draw position=4.25,
] table[row sep=\\, y index=0] {
data \\
57	\\
31	\\
44	\\
61	\\
30	\\
41	\\
36	\\
105	\\
68	\\
101	\\
63	\\
100	\\
51	\\
26	\\
29	\\
33	\\
30	\\
52	\\
34	\\
34	\\
38	\\
57	\\
33	\\
29	\\
47	\\
33	\\
19	\\
57	\\
86	\\
};

% k = 90
\addplot[
    boxplot,
    sumrate,
    boxplot/draw position=4.75,
] table[row sep=\\, y index=0] {
data \\
68	\\
46	\\
132	\\
99	\\
140	\\
84	\\
77	\\
122	\\
81	\\
66	\\
85	\\
120	\\
130	\\
135	\\
158	\\
123	\\
144	\\
161	\\
91	\\
88	\\
161	\\
106	\\
51	\\
122	\\
157	\\
52	\\
66	\\
114	\\
54	\\
};
\addplot[
    boxplot,
    power,
    boxplot/draw position=5.25,
] table[row sep=\\, y index=0] {
data \\
35	\\
42	\\
38	\\
71	\\
34	\\
25	\\
63	\\
25	\\
55	\\
44	\\
39	\\
44	\\
29	\\
24	\\
32	\\
30	\\
55	\\
100	\\
49	\\
48	\\
47	\\
90	\\
45	\\
75	\\
53	\\
49	\\
40	\\
27	\\
22	\\
};

% k = 100
\addplot[
    boxplot,
    sumrate,
    boxplot/draw position=5.75,
] table[row sep=\\, y index=0] {
data \\
81	\\
63	\\
81	\\
98	\\
144	\\
89	\\
90	\\
104	\\
128	\\
137	\\
141	\\
104	\\
96	\\
67	\\
129	\\
140	\\
262	\\
130	\\
193	\\
91	\\
134	\\
102	\\
97	\\
33	\\
119	\\
61	\\
73	\\
136	\\
116	\\
};
\addplot[
    boxplot,
    power,
    boxplot/draw position=6.25,
] table[row sep=\\, y index=0] {
data \\
29	\\
30	\\
50	\\
35	\\
98	\\
36	\\
62	\\
33	\\
29	\\
38	\\
36	\\
45	\\
27	\\
33	\\
38	\\
88	\\
33	\\
20	\\
52	\\
24	\\
26	\\
50	\\
72	\\
33	\\
41	\\
41	\\
23	\\
34	\\
41	\\
};

% k = 110
\addplot[
    boxplot,
    sumrate,
    boxplot/draw position=6.75,
] table[row sep=\\, y index=0] {
data \\
45	\\
126	\\
119	\\
144	\\
88	\\
101	\\
119	\\
104	\\
214	\\
96	\\
72	\\
109	\\
90	\\
86	\\
91	\\
94	\\
123	\\
120	\\
90	\\
183	\\
76	\\
142	\\
39	\\
61	\\
71	\\
84	\\
64	\\
169	\\
155	\\
};
\addplot[
    boxplot,
    power,
    boxplot/draw position=7.25,
] table[row sep=\\, y index=0] {
data \\
35	\\
60	\\
29	\\
101	\\
126	\\
36	\\
46	\\
36	\\
37	\\
83	\\
35	\\
68	\\
19	\\
63	\\
42	\\
34	\\
44	\\
28	\\
41	\\
24	\\
39	\\
44	\\
50	\\
47	\\
59	\\
60	\\
21	\\
50	\\
33	\\
};

% k = 120
\addplot[
    boxplot,
    sumrate,
    boxplot/draw position=7.75,
] table[row sep=\\, y index=0] {
data \\
223	\\
161	\\
89	\\
60	\\
67	\\
56	\\
45	\\
165	\\
40	\\
44	\\
19	\\
25	\\
138	\\
96	\\
88	\\
94	\\
91	\\
100	\\
103	\\
57	\\
46	\\
60	\\
127	\\
40	\\
79	\\
142	\\
49	\\
150	\\
41	\\
};
\addplot[
    boxplot,
    power,
    boxplot/draw position=8.25,
] table[row sep=\\, y index=0] {
data \\
41	\\
30	\\
68	\\
51	\\
43	\\
33	\\
56	\\
63	\\
25	\\
204	\\
42	\\
53	\\
48	\\
54	\\
52	\\
38	\\
53	\\
43	\\
49	\\
77	\\
30	\\
33	\\
58	\\
35	\\
56	\\
31	\\
32	\\
41	\\
28	\\
};

% k = 130
\addplot[
    boxplot,
    sumrate,
    boxplot/draw position=8.75,
] table[row sep=\\, y index=0] {
data \\
114	\\
110	\\
189	\\
117	\\
56	\\
54	\\
78	\\
46	\\
116	\\
39	\\
79	\\
88	\\
36	\\
82	\\
52	\\
62	\\
60	\\
111	\\
84	\\
85	\\
98	\\
110	\\
119	\\
66	\\
46	\\
43	\\
80	\\
138	\\
164	\\
};
\addplot[
    boxplot,
    power,
    boxplot/draw position=9.25,
] table[row sep=\\, y index=0] {
data \\
32	\\
80	\\
46	\\
80	\\
58	\\
50	\\
51	\\
45	\\
89	\\
32	\\
47	\\
46	\\
44	\\
92	\\
81	\\
92	\\
40	\\
63	\\
71	\\
44	\\
53	\\
40	\\
65	\\
58	\\
53	\\
20	\\
64	\\
67	\\
35	\\
};

% k = 140
\addplot[
    boxplot,
    sumrate,
    boxplot/draw position=9.75,
] table[row sep=\\, y index=0] {
data \\
140	\\
94	\\
102	\\
46	\\
43	\\
77	\\
111	\\
86	\\
87	\\
39	\\
44	\\
28	\\
46	\\
71	\\
54	\\
60	\\
125	\\
25	\\
92	\\
157	\\
70	\\
146	\\
49	\\
68	\\
87	\\
64	\\
77	\\
130	\\
138	\\
};
\addplot[
    boxplot,
    power,
    boxplot/draw position=10.25,
] table[row sep=\\, y index=0] {
data \\
53	\\
61	\\
106	\\
107	\\
75	\\
122	\\
54	\\
63	\\
45	\\
78	\\
66	\\
51	\\
49	\\
68	\\
32	\\
46	\\
83	\\
63	\\
53	\\
85	\\
33	\\
35	\\
39	\\
53	\\
59	\\
41	\\
79	\\
52	\\
76	\\
};

% k = 150
\addplot[
    boxplot,
    sumrate,
    boxplot/draw position=10.75,
] table[row sep=\\, y index=0] {
data \\
98	\\
48	\\
68	\\
50	\\
23	\\
59	\\
75	\\
73	\\
95	\\
114	\\
37	\\
38	\\
38	\\
71	\\
107	\\
74	\\
39	\\
56	\\
51	\\
47	\\
91	\\
79	\\
70	\\
46	\\
59	\\
69	\\
36	\\
53	\\
67	\\
};
\addplot[
    boxplot,
    power,
    boxplot/draw position=11.25,
] table[row sep=\\, y index=0] {
data \\
49	\\
87	\\
85	\\
37	\\
61	\\
80	\\
114	\\
50	\\
46	\\
73	\\
79	\\
79	\\
77	\\
55	\\
75	\\
86	\\
65	\\
57	\\
68	\\
92	\\
65	\\
69	\\
64	\\
72	\\
72	\\
78	\\
48	\\
56	\\
};

\end{axis}
\end{tikzpicture}%
  }
  \caption{\omini{}}
\end{subfigure}
\hfill
\begin{subfigure}[t]{0.48\textwidth}
  \centering
  \resizebox{\linewidth}{!}{%
    \begin{tikzpicture}
\begin{axis}[
    width=8cm,
    height=4.6cm,
    ymin=0,
    ymax=500,
    xlabel={$K$},
    ylabel={Execution time (s)},
    boxplot/draw direction=y,
    boxplot={
        box extend=0.2
    },
    xtick={1,2,3, 4, 5, 6, 7, 8, 9},
    xticklabels={80, 90, 100, 110, 120, 130, 140, 150, 160},
    every axis plot/.append style={thick},
    sumrate/.style={draw=blue, fill=blue!20, solid},
    power/.style={draw=red, fill=red!20, solid},
    legend columns=2,
    legend style={
        at={(axis description cs:1,1.10)},
        anchor=south east,
        draw=black,
        font=\scriptsize,
        /tikz/overlay,
        /tikz/every even column/.append style={column sep=5pt},
    },
    legend image code/.code={
        \draw[#1] (0cm,-0.05cm) rectangle (0.18cm,0.12cm);
    },
]

% k = 80
\addplot[
    boxplot,
    sumrate,
    boxplot/draw position=0.75,
] table[row sep=\\, y index=0] {
data \\
86	\\
87	\\
263	\\
172	\\
128	\\
114	\\
134	\\
106	\\
159	\\
110	\\
75	\\
95	\\
135	\\
147	\\
99	\\
89	\\
125	\\
92	\\
232	\\
146	\\
92	\\
103	\\
98	\\
106	\\
119	\\
111	\\
115	\\
71	\\
};
\addlegendentry{Sum-rate}
\addplot[
    boxplot,
    power,
    boxplot/draw position=1.25,
] table[row sep=\\, y index=0] {
data \\
129	\\
110	\\
119	\\
143	\\
150	\\
164	\\
150	\\
180	\\
142	\\
146	\\
183	\\
106	\\
144	\\
140	\\
162	\\
159	\\
164	\\
124	\\
100	\\
172	\\
125	\\
107	\\
114	\\
144	\\
146	\\
131	\\
136	\\
85	\\
};
\addlegendentry{Power efficiency}

% k = 90
\addplot[
    boxplot,
    sumrate,
    boxplot/draw position=1.75,
] table[row sep=\\, y index=0] {
data \\
121	\\
101	\\
113	\\
118	\\
121	\\
65	\\
124	\\
130	\\
185	\\
110	\\
126	\\
145	\\
92	\\
98	\\
99	\\
155	\\
90	\\
104	\\
107	\\
299	\\
99	\\
97	\\
105	\\
94	\\
92	\\
97	\\
119	\\
146	\\
};
\addplot[
    boxplot,
    power,
    boxplot/draw position=2.25,
] table[row sep=\\, y index=0] {
data \\
134	\\
153	\\
113	\\
129	\\
117	\\
144	\\
150	\\
129	\\
129	\\
110	\\
111	\\
120	\\
105	\\
129	\\
149	\\
119	\\
117	\\
153	\\
166	\\
108	\\
137	\\
153	\\
134	\\
128	\\
134	\\
146	\\
134	\\
121	\\
};

% k = 100
\addplot[
    boxplot,
    sumrate,
    boxplot/draw position=2.75,
] table[row sep=\\, y index=0] {
data \\
132	\\
84	\\
203	\\
173	\\
115	\\
107	\\
104	\\
109	\\
108	\\
109	\\
153	\\
177	\\
89	\\
91	\\
122	\\
136	\\
218	\\
83	\\
99	\\
124	\\
119	\\
95	\\
113	\\
112	\\
228	\\
108	\\
125	\\
};
\addplot[
    boxplot,
    power,
    boxplot/draw position=3.25,
] table[row sep=\\, y index=0] {
data \\
137	\\
130	\\
103	\\
119	\\
98	\\
108	\\
128	\\
106	\\
166	\\
137	\\
117	\\
120	\\
117	\\
129	\\
133	\\
170	\\
112	\\
125	\\
123	\\
123	\\
146	\\
138	\\
109	\\
136	\\
152	\\
138	\\
137	\\
90	\\
135	\\
};

% k = 110
\addplot[
    boxplot,
    sumrate,
    boxplot/draw position=3.75,
] table[row sep=\\, y index=0] {
data \\
116	\\
120	\\
113	\\
146	\\
135	\\
98	\\
144	\\
119	\\
139	\\
140	\\
103	\\
247	\\
176	\\
126	\\
285	\\
104	\\
233	\\
106	\\
140	\\
93	\\
132	\\
141	\\
134	\\
172	\\
140	\\
141	\\
140	\\
134	\\
};
\addplot[
    boxplot,
    power,
    boxplot/draw position=4.25,
] table[row sep=\\, y index=0] {
data \\
101	\\
145	\\
118	\\
136	\\
139	\\
150	\\
123	\\
115	\\
126	\\
153	\\
145	\\
142	\\
128	\\
124	\\
152	\\
113	\\
134	\\
142	\\
146	\\
153	\\
145	\\
145	\\
132	\\
124	\\
166	\\
160	\\
120	\\
148	\\
135	\\
};

% k = 120
\addplot[
    boxplot,
    sumrate,
    boxplot/draw position=4.75,
] table[row sep=\\, y index=0] {
data \\
253	\\
241	\\
140	\\
135	\\
151	\\
164	\\
146	\\
223	\\
108	\\
165	\\
125	\\
85	\\
113	\\
155	\\
155	\\
153	\\
173	\\
295	\\
195	\\
151	\\
324	\\
160	\\
99	\\
150	\\
124	\\
163	\\
151	\\
117	\\
};
\addplot[
    boxplot,
    power,
    boxplot/draw position=5.25,
] table[row sep=\\, y index=0] {
data \\
132	\\
135	\\
147	\\
118	\\
153	\\
137	\\
136	\\
132	\\
140	\\
113	\\
136	\\
137	\\
119	\\
158	\\
170	\\
118	\\
165	\\
110	\\
113	\\
143	\\
113	\\
133	\\
124	\\
131	\\
113	\\
168	\\
123	\\
161	\\
123	\\
};

% k = 130
\addplot[
    boxplot,
    sumrate,
    boxplot/draw position=5.75,
] table[row sep=\\, y index=0] {
data \\
158	\\
123	\\
106	\\
136	\\
286	\\
105	\\
142	\\
209	\\
143	\\
486	\\
187	\\
124	\\
359	\\
136	\\
116	\\
144	\\
121	\\
138	\\
150	\\
137	\\
147	\\
173	\\
184	\\
146	\\
113	\\
133	\\
161	\\
};
\addplot[
    boxplot,
    power,
    boxplot/draw position=6.25,
] table[row sep=\\, y index=0] {
data \\
129	\\
160	\\
111	\\
141	\\
109	\\
113	\\
132	\\
138	\\
149	\\
113	\\
120	\\
141	\\
188	\\
154	\\
111	\\
122	\\
120	\\
154	\\
100	\\
125	\\
155	\\
122	\\
126	\\
122	\\
133	\\
114	\\
152	\\
119	\\
142	\\
};

% k = 140
\addplot[
    boxplot,
    sumrate,
    boxplot/draw position=6.75,
] table[row sep=\\, y index=0] {
data \\
142	\\
296	\\
134	\\
461	\\
417	\\
138	\\
220	\\
158	\\
153	\\
129	\\
129	\\
262	\\
181	\\
143	\\
127	\\
141	\\
139	\\
203	\\
187	\\
132	\\
108	\\
153	\\
118	\\
127	\\
146	\\
153	\\
163	\\
};
\addplot[
    boxplot,
    power,
    boxplot/draw position=7.25,
] table[row sep=\\, y index=0] {
data \\
175	\\
117	\\
142	\\
133	\\
114	\\
136	\\
165	\\
199	\\
137	\\
128	\\
141	\\
165	\\
152	\\
139	\\
134	\\
158	\\
113	\\
148	\\
134	\\
127	\\
153	\\
161	\\
147	\\
112	\\
155	\\
132	\\
150	\\
141	\\
129	\\
};

% k = 150
\addplot[
    boxplot,
    sumrate,
    boxplot/draw position=7.75,
] table[row sep=\\, y index=0] {
data \\
105	\\
109	\\
123	\\
110	\\
301	\\
176	\\
270	\\
144	\\
147	\\
125	\\
131	\\
159	\\
133	\\
124	\\
119	\\
190	\\
120	\\
493	\\
278	\\
102	\\
101	\\
114	\\
136	\\
135	\\
116	\\
164	\\
134	\\
133	\\
};
\addplot[
    boxplot,
    power,
    boxplot/draw position=8.25,
] table[row sep=\\, y index=0] {
data \\
116	\\
121	\\
103	\\
139	\\
161	\\
153	\\
155	\\
149	\\
139	\\
122	\\
91	\\
118	\\
146	\\
145	\\
116	\\
147	\\
149	\\
130	\\
134	\\
154	\\
127	\\
130	\\
158	\\
141	\\
121	\\
165	\\
118	\\
151	\\
};

% k = 160
\addplot[
    boxplot,
    sumrate,
    boxplot/draw position=8.75,
] table[row sep=\\, y index=0] {
data \\
166	\\
149	\\
136	\\
139	\\
75	\\
117	\\
151	\\
160	\\
125	\\
141	\\
184	\\
150	\\
126	\\
144	\\
154	\\
156	\\
215	\\
153	\\
162	\\
211	\\
155	\\
128	\\
173	\\
150	\\
154	\\
124	\\
155	\\
165	\\
133	\\
};
\addplot[
    boxplot,
    power,
    boxplot/draw position=9.25,
] table[row sep=\\, y index=0] {
data \\
139	\\
154	\\
180	\\
137	\\
170	\\
135	\\
153	\\
160	\\
132	\\
111	\\
113	\\
132	\\
151	\\
139	\\
128	\\
108	\\
144	\\
128	\\
164	\\
149	\\
154	\\
172	\\
113	\\
116	\\
136	\\
132	\\
117	\\
115	\\
234	\\
143	\\
158	\\
139	\\
116	\\
};
\end{axis}
\end{tikzpicture}%
  }
  \caption{\gem{}}
\end{subfigure}

\caption{Execution time distributions for attempts with different
values of $K$. For \omini{} (a), shorter runtimes are observed
in the more complex fractional programming task. In contrast,
\gem{} (b) exhibits more consistent execution
times across both problem types.}
\label{fig:k-time-boxplot}
\end{figure*}

Figure~\ref{fig:k-time-boxplot} shows the runtime distributions for both models executing the two power-allocation tasks. The LLM response time remains relatively stable across the evaluated values of $K$, although the two models exhibit different task-dependent latency patterns. For \omini{}, fractional programming often has a lower median latency and smaller variance than sum-rate maximization. For \gem{}, the difference between the two tasks is less consistent as $K$ increases.

\section{Discussions}
\label{sec:discussions}
% \subsection{What Makes Continuous Optimization Difficult for LLMs?}
\subsection{Main Source of Difficulty for LLMs}

The results suggest that it is difficult for LLMs to satisfy all requirements simultaneously. Among all requirements, numerical precision is a key challenge. In the two power-allocation tasks, model outputs often follow reasonable allocation patterns, but violate budget requirements by a margin. In the projection task, candidate solutions often fail to fulfill the equality constraint under strict tolerance.

High-dimensional counting and structured-output generation are additional limitations. At larger $K$, the dominant failure mode is the incorrect number of decision variables. Format violations also prevent the model from producing valid outputs for following steps. Besides the infamous difficulty of LLMs with counting studied in previous work, e.g., \cite{counting1, counting2}, these failures also suggest that LLMs face challenges to abide by strict output specifications, when context windows are large and the output is long.

Finally, optimization difficulty depends on the mathematical structure of the task. Sum-rate maximization, which describes a well-known classical convex programming task, produces the strongest overall results. Energy-efficiency maximization, which has a more demanding fractional objective, shows lower optimality rates and earlier failures. The projection task shows a different limitation, as it includes a strict equality constraint, making feasibility more sensitive to numerical precision. 

\subsection{Understanding LLM Optimization Capabilities}

The empirical results indicate that the evaluated models capture useful high-level optimization structure. In sum-rate maximization, LLM-generated allocations often resemble the deterministic water-filling solution. However, a small budget violation may still lead to invalid outputs, as shown in \comparisonlocation{}. In the projection task, the high optimality rate among valid outputs indicates that the model can generate solutions close to the constrained optimum as long as feasibility is achieved. However, LLMs cannot reliably perform the exact arithmetic and strict constraint enforcement required by numerical optimization. Structured feedback helps models correct some formatting, counting, and feasibility errors and improve objective quality. Nevertheless, iterative refinement does not guarantee success, particularly at larger dimensions.

Meanwhile, although \omini{} and \gem{} exhibit different optimization capabilities, their results reveal similar overall trends. \gem{} shows a larger operating range, but still suffers from failures at larger dimensions. This suggests that stronger models can extend the usable range of direct LLM optimization, while reliable large-scale applications still require additional mechanisms for verification and repair.

\subsection{Implications for LLM-Based Optimization}

These findings have several implications for the design of LLM-based optimization systems. First, prompts should state output and feasibility requirements explicitly. They should specify the expected output length, and numerical format. All constraints should be given in a \textit{machine-checkable} form. Shorter prompts, clearer instructions, and strategic placement of key requirements may provide further improvements. When supported by the model interface, output schemas, JSON-mode generation, function calling, or constrained decoding may further improve reliability.

% Second, 
The results further suggest that the LLMs are more useful as components of hybrid optimization systems rather than as standalone solvers. An LLM can interpret a natural-language specification and propose initial solutions, followed by a deterministic solver or numerical verification step to ensure feasibility and optimality. This division of responsibility combines the flexibility of LLMs with the reliability and guarantees of conventional optimization methods.

The findings also imply that rather than focusing on the final objective values, more metrics should be used to evaluate LLM optimization performance in future benchmarks. Validity, failure mode distribution, format compliance and the magnitude of constraint violation can all be used for characterization.

We see these findings as a starting point for future research on LLM-based optimization. Future work should address the execution errors that limit LLM-based optimization on large numerical problems. One direction is to enforce structured outputs, such as using predefined JSON schemas. Better prompts and context-management methods may also improve reliability. Another option is to decompose the problem so that the model does not need to generate a long numerical vector at once. LLMs could also be combined with conventional solvers. In such a system, the LLM would handle high-level reasoning, while the solver would ensure numerical accuracy and feasibility. These methods should be tested on more optimization problems, larger constraint sets, and real engineering applications. Future studies should also reduce inference time and the cost of iterative refinement. Together, these improvements could make LLMs more useful and reliable within practical optimization pipelines.

\section{Conclusions}
\label{sec:conclusions}
This paper presented a systematic evaluation of LLMs for numerical optimization. We benchmarked LLM-based optimization loops %\texttt{o4-mini} and \texttt{gemini-3-flash-preview} 
against deterministic reference solvers on three representative problems: sum-rate maximization, energy-efficiency maximization, and constrained minimum-distance projection. %The evaluation considered solution validity, objective quality relative to deterministic reference solutions, failure modes, scaling with problem dimension, and runtime.
The results show that the performance of LLM-based solvers depends strongly on the optimization family and problem dimension. For small and moderate problem instances, models can produce valid and near-optimal solutions, while at large dimensions, the models increasingly exhibit errors. %, output-format violations, and numerical-precision errors. These failures decrease the likelihood of generating valid solutions. 
Sum-rate maximization supported the widest range of valid and high-quality outputs, whereas energy-efficiency maximization exhibited lower optimality rates and earlier reliability failures. In the projection task, valid outputs were frequently near-optimal, but feasibility was sensitive to the equality-constraint tolerance.
% We also evaluated 
Iterative refinement %as an optimization protocol. Structured validation feedback allowed the models to revise their outputs. Compared with the single-turn setting, this process 
can improve the validity and optimality of LLM-based solver and extends its scalability. % increased the range of dimensions for which the models could produce feasible solutions. However,
Nonetheless, it cannot eliminate persistent %counting, formatting, or numerical 
failures and increases end-to-end latency. % by requiring additional model calls.

% Overall, 
The results suggest that LLMs are best viewed as flexible candidate-solution generators rather than replacements for deterministic solvers. Their ability to interpret optimization problems and propose competitive solutions can be useful when problem formulation, solver selection, or human interaction is difficult to automate. Applications that require strict feasibility, reproducibility, or low latency should consider a hybrid approach. In such systems, an LLM generates candidate solutions, while deterministic algorithms validate and refine them. Future work can investigate constrained output generation, solver-in-the-loop refinement, and problem decompositions for high-dimensional tasks. %, and broader benchmarks that measure feasibility, scaling behaviour, failure modes, and computational cost.

\section*{Acknowledgment}
The authors used LLMs solely for grammar correction and language polishing throughout the manuscript. All suggested revisions were reviewed by the authors, who take full responsibility for the content.

\appendices
\section{LLM Prompt and Results}
Below, we present a sample prompt and the results generated by the LLM for sum-rate maximization task. 
\subsection{Initial Prompt for Sum-Rate Maximization}
\label{app:waterfilling_initial_prompt}
\begin{lstlisting}
    You are solving a problem by choosing nonnegative number x_1,...,x_K (sum capped) to maximize the total rate R = sum(log2(L_i)) where L_i = 1 + x_i * g_i / N_i.
    
    Concepts to use:
    * For fixed others, increasing x_i increases log2(L_i).
    * Concavity: The extra benefit from adding to a large L_i is small than from adding to a small L_i.
    * Balancing: Increasing a small L_i while decreasing a large L_j tends to increase total R.
    * Implication: For a total sum(x_i) with fixed maximum, more balanced levels L_i tend to give higher total rate than very uneven ones.
    
    Example:
    P = 15, G = [3,1,9], N = [1, 1, 1]
    x = [0,0,15] -> L = [1,1,136] -> lower R
    x = [4, 10, 1] -> L = [13, 11, 10] -> higher R
    
    Task:
    K = 10
    P = sum(x_i) <= 3
    
    g_1=3.6409156481810863, N_1=1.0
    g_2=0.11952989450819125, N_2=1.0
    g_3=1.1623212806322352, N_3=1.0
    g_4=5.496000867022677, N_4=1.0
    g_5=0.9760133249930699, N_5=1.0
    g_6=0.4715046846895365, N_6=1.0
    g_7=1.5327662965629543, N_7=1.0
    g_8=0.14916894342800793, N_8=1.0
    g_9=4.273319004236757, N_9=1.0
    g_10=1.643596707263682, N_10=1.0
    
    
    Reply format:
    Return a list of K nonnegative numbers summing to P, wrapped in square brackets, e.g., [54, 33, 33].
    
    Now output the optimal list [x_1, x_2, ..., x_K].
    Output only the list, no explanations or text.
\end{lstlisting}
\subsection{LLM Result and Deterministic Solution Comparison}
%\subsection{Sum-Rate Maximization}
\label{app:sum_rate_solution_comparison}
The following results in Table \ref{tab:alloc_example} are two overbudget solution produced by LLM models (o4-mini and gemini-3-flash-preview) on scenarios with $K=150$ channels and $P_{\text{total}}=60$, compared with their respective deterministic optimal solution.
\begin{table*}[!t]
\centering
\caption{Sample overbudget allocations comparing raw decisions from \omini{} and \gem{} to a deterministic optimum ($K=150$, $P_{\text{total}}=60$). While LLMs grasp the water-filling distribution pattern intuitively, poor arithmetic precision inherently breaks the total budget.}
\label{tab:alloc_example}
\renewcommand{\arraystretch}{1.1}
\setlength{\tabcolsep}{4pt}
\small
\begin{tabular}{@{}l r r r r r r r r@{}}
\toprule
 & $\sum{x_i}$ & $x_1$ & $x_2$ & $x_3$ & $x_4$ & $x_5$ & $x_6$ & $\ldots$ \\
\midrule
\omini{} & 60.4482 & 0& 0.595& 0.0835& 0& 0.6618& 0.7147& $\ldots$ \\
optimal  & 60.0 & 0.0& 0.5905688075& 0.07903040838& 0.0& 0.6574058524& 0.7101931468& $\ldots$ \\
\midrule\midrule
\gem{} & 60.0105 & 0 & 0.6804 & 0 & 0.8104 & 0.3233 & 0  & $\ldots$ \\
optimal  & 60.0 & 0& 0.6803830239& 0& 0.8103994294& 0.3233392392& 0 & $\ldots$\\
\bottomrule
\end{tabular}
\end{table*}

\section{LLM Reasoning Trace}
\label{app:reasoning_traces}
The following are representative reasoning traces (thinking summaries) produced by \gem{} when solving optimization problems. These traces are extracted from the model's internal chain-of-thought and illustrate how the LLM approaches each problem.

\subsection{Sum-Rate Maximization}
\label{app:sumrate_thinking}
The following trace shows the model solving a sum-rate maximization instance. Notably, the model derives the water-filling strategy \textit{without} identifying the problem as one being solved by ``water-filling''.

\begin{lstlisting}
Alright, let's break this down. My objective is to maximize the sum
of logarithms of levels, each level being a combination of a fixed
component and a resource allocation multiplied by a constant. The
key here is the concavity of the logarithm, which pushes me towards
balancing these levels as much as possible. I've got a budget
constraint - I can only allocate a total of 6000 units of the
resource. There are 50 different "floors", and an increment of 0.05
can be added to the base for each floor. The f_i values are provided.

My initial approach is to try and make each level roughly equal. I
can think about this by setting a target level, L*. I can try to
solve for x_i, the resource allocation to each floor, by setting
f_i + x_i * delta approx L*. The smallest base and Q=6000.0 are
provided.

Calculating: I first determine the target level L*. If I could use
fractions of resource quanta, x_i = (L* - f_i) / delta, setting
f_i + x_i * delta = L*. I can sum over all i to solve for L*, and
determine how much resource to allocate to each of the 50 "floors".
But the prompt has integer values, so I have to come up with whole
units.

I see a pattern, where the final allocation will be
x_i = 20 * (10.68 - f_i). This yields non-integer values for x_i.
I will calculate x_i = (L* - f_i) / delta, which might lead to
fractional numbers.

I take the sum of the f_i values to be 234. I find the target level
L* = (Q * delta + sum(f_i)) / |S|, which comes out as 10.68. The
set S includes all 50 floors. I calculate a value of x_i.
I calculate Q * delta = 300, so my formula is
L* = (300 + 234) / 50 = 534 / 50 = 10.68.

I will have to check the fractional value of x_i to ensure that
sum(x_i) is at most 6000. I realize all the x_i have the form
213.6 - 20 * f_i, and all fractional parts will have 0.6. I see a
solution! I realize that I have to round all the x_i values to
integers, and to make it fit with the constraint that the sum equals
6000, 30 must be rounded up, and 20 must be rounded down.

I see the sum of the allocated resource is 6000.
L_i = f_i + x_i * 0.05. So when the rounding occurs, the values
are almost the same.

I have the set of x_i values. All non-negative integers.
L_i values are f_i + x_i * delta. I need the 30 round ups and the
20 round downs. Sum of allocation equals 6000.
The L_i values are almost identical, and very balanced.
I verify that my logic is correct.
I see the allocation must be integers.
I've chosen 30 indices to round up.
I've chosen 20 indices to round down.
I recalculate. The total sum of allocation is exactly 6000, as
required.

The rounding strategy doesn't impact the overall result.
Therefore, the allocations have been decided.
I choose to assign the x_i to ensure L_i are as balanced as
possible. I double-check. I can show that sum(log2(L_i)) will be
very large.
\end{lstlisting}

\subsection{Minimum-Distance Projection}
\label{app:projection_thinking}
The following trace shows the model solving a projection problem instance with $K=40$ variables, one equality constraint, and one inequality constraint. The model applies KKT conditions and Lagrangian analysis from first principles to derive the solution structure. The trace was truncated by the model's reasoning token limit before the numerical computation was completed. However, the reasoning up to that point demonstrates correct formulation and case analysis.

\begin{lstlisting}
Okay, here's the deal. I'm faced with a minimization problem. The
goal is to make f(x) = 1/2 ||x - y||^2 as small as possible,
effectively finding the point x closest to a fixed point y. However,
there are rules. I'm dealing with a vector x in 40 dimensions
(K=40, y is a 40-dim vector, x is the variable I'm optimizing
over), so that the objective is to find a vector x, subject to a
few constraints.

First, all components of x must be non-negative: x_i >= 0. This is
straightforward. Second, there's an equality constraint: Ax + b = 0.
Think of A as a row vector (1x40 matrix), and b is just a number.
It defines a hyperplane. My solution x must lie on this plane, since
I want Ax=-b, given A_i (40 element vector) and b (the intercept),
and x_i which are the elements of x (40 element vector). Finally,
I've got an inequality constraint: Ux <= d. Here U is another row
vector (1x40 matrix), d a scalar. So, I need to keep x "below"
this hyperplane too.

To handle these constraints formally, I'm using the Lagrangian
method. I introduce Lagrange multipliers: lambda for the equality
constraint, mu >= 0 for the inequality, and nu_i >= 0 for the
non-negativity constraints. The Lagrangian L(x, lambda, mu, nu) is
formed by combining the original objective function with the
constraints, each multiplied by its corresponding Lagrange
multiplier.

Applying the Karush-Kuhn-Tucker (KKT) conditions is the next step.
I'm differentiating the Lagrangian with respect to x and setting it
to zero, which gives me:
x_i = y_i - lambda A_i - mu U_i + nu_i.
Furthermore, I recognize that x_i >= 0, nu_i >= 0 and
x_i * nu_i = 0. This implies
x_i = max(0, y_i - lambda A_i - mu U_i).
I also have to obey that Ax + b = 0, and the inequality constraint
U x - d <= 0, mu >= 0, mu (U x - d) = 0.

Now, I split into cases. The most obvious one to consider is when
mu = 0. That is, the inequality constraint is inactive, and not
binding. The inequality constraint is non-binding if the solution x
satisfies Ux < d. In this scenario,
x_i = max(0, y_i - lambda A_i).
The main goal now is to find lambda so that the equality constraint,
Sum A_i x_i + b = 0 holds. Or solving for lambda in
g(lambda) = Sum A_i max(0, y_i - lambda A_i) + b = 0.
The next, and critical, question is whether to actually choose mu or
not. Let's see if this choice is the right choice. Finally, I will
check whether the solution x satisfies the inequality constraint,
and if it satisfies U x <= d.

If that doesn't work (which, I suspect, it won't), the next case,
mu > 0, becomes active. Now, the inequality is binding. This means
that my solution x must satisfy the inequality with equality:
Ux = d. Here, x_i = max(0, y_i - lambda A_i - mu U_i). Now, I have
two unknowns: lambda and mu. I need to solve two equations:
Sum A_i x_i + b = 0 and Sum U_i x_i - d = 0.

Ok, now, I'm given the specific values: A, b, U, and y. And that's
where things get interesting. I started by checking mu = 0. I
computed g(lambda) = Sum A_i max(0, y_i - lambda A_i) + b = 0,
setting b = -2.5285. Using some algebra, I found
\end{lstlisting}
\reviewchange{\textit{[Trace truncated at the model's reasoning token limit.]}}

\bibliographystyle{IEEEtran}
\bibliography{references}
\end{document}